\documentclass{article}

\PassOptionsToPackage{numbers, compress}{natbib}

\usepackage[preprint]{neurips_2026}
\usepackage{svg}      
\usepackage{xstring}  
\usepackage{amsmath}
\usepackage{wrapfig}
\usepackage{caption}  
\usepackage{subcaption}
\usepackage{amssymb}

\usepackage[utf8]{inputenc} 
\usepackage[T1]{fontenc}    
\usepackage{hyperref}       
\usepackage{url}            
\usepackage{booktabs}       
\usepackage{amsfonts}       
\usepackage{nicefrac}       
\usepackage{microtype}      
\usepackage[table]{xcolor}  
\usepackage{multirow}
\usepackage{adjustbox}
\usepackage{graphicx}
\usepackage{array}
\usepackage{makecell}
\usepackage{fontawesome5} 
\usepackage{float}

\usepackage{times}
\usepackage{latexsym}
\usepackage{inconsolata}
\usepackage{tcolorbox}
\tcbuselibrary{skins,breakable,xparse}

\newtcolorbox{userbox}[2][]{
  enhanced,
  breakable,
  colback=blue!3,
  colframe=blue!60!black,
  coltitle=white,
  fonttitle=\bfseries\small\ttfamily\color{white},
  fontupper=\ttfamily\footnotesize,
  boxrule=0.8pt,
  title={#2},
  title style={fill=blue!60!black},
  left=5mm, right=5mm,
  top=3mm, bottom=3mm,
  arc=2mm,
  borderline west={3pt}{0pt}{blue!70},
  boxsep=2pt,
  #1
}

\title{The Last Visible Pixel: Probing Fine-Scale Perception in Vision-Language Models}

\author{
Lujun Li\textsuperscript{1},
Lama Sleem\textsuperscript{1},
Niccolo' Gentile\textsuperscript{2},
Yangjie Xu\textsuperscript{1},\\
\textbf{Yewei Song\textsuperscript{1},
Wenbo Wu\textsuperscript{3},
Radu State\textsuperscript{1}} \\
\textsuperscript{1}University of Luxembourg,
\textsuperscript{2}Foyer S.A.,
\textsuperscript{3}Université Paris-Saclay
}

\begin{document}

\maketitle

\begin{abstract}
Recent vision-language models (VLMs) excel at multimodal understanding and reasoning, yet their fine-grained visual perception remains underexplored. A natural extension of ``How many r are there in \textit{Strawberry}?'' asks: how small a visual pattern can a VLM reliably perceive? As such, we introduce FineSightBench, a new benchmark that systematically probes this limit by separating perception tasks (pixel-level recognition of letters, shapes, objects) from reasoning tasks (spatial reasoning, counting, ordering over small targets) across controlled scales of 4--48px. Through comprehensive experiments and detailed failure mode analysis on state-of-the-art models, we reveal a sharp dissociation: perception saturates around 12px, while reasoning remains limited even at larger scales, with persistent numeracy and sequence errors. These findings expose fundamental deficiencies in VLMs' fine-scale visual reasoning that demand more rigorous evaluation.
\end{abstract}

\section{Introduction}

Vision-language models (VLMs,~\cite{bai2023qwenvlversatilevisionlanguagemodel,li2025surveystateartlarge}) have recently emerged as a prominent paradigm for bridging visual signals and natural language, enabling a wide range of applications. Despite the substantial improvements reported on standard benchmark datasets, these evaluations offer only limited insight into a fundamental capability that is crucial for many real-world applications, namely, the fine-grained perception of small-scale, subtle, and spatially localized visual details. Although prior work has highlighted the visual limitations of VLMs and, through intervention-based causal analysis, demonstrated that the increased difficulty of perceiving small objects is not only an observed correlation but a causal factor underlying performance degradation~\cite{zhang2025mllmsknowlooktrainingfree}, the notion of how small an object is remains insufficiently quantified, especially in terms of its pixel-level scale. In practice, these models are often assumed to operate on patch-level visual tokens, which raises fundamental uncertainty regarding the minimum object size they can reliably perceive~\cite{DBLP:journals/corr/abs-2010-11929,feng2025visionlanguagemodelobjectdetection}. Moreover, accurately reporting a visual detail is not equivalent to using it correctly for accurate reasoning and faithful description: a VLM may occasionally detect a subtle cue, yet still fail to reason over it consistently or express it in a verifiable manner.

To make these capabilities measurable, we introduce \textbf{FineSightBench}, designed to evaluate fine-scale visual ability of VLMs. This consists of two complementary components. The first, \textbf{Perception}, targets pixel-level visual perception by varying the maximum object dimensions (width and height) and the subtlety of small objects, thereby probing the detection limits of state-of-the-art VLMs. The second, \textbf{Reasoning}, focuses on evaluating models' sensitivity to perform spatially precise, detail-intensive reasoning at different pixel scales. More specifically, in this paper, fine-scale perception in VLMs is investigated through the lens of two progressive perspectives: (i) \textbf{seeing fine-grained visual signals}, and (ii) \textbf{seeing, reasoning and describing correctly on the basis of those signals}~\cite{berman2025vlmstunnelvisionevaluating}.

\begin{figure}[H]
  \centering
  \includegraphics[width=\linewidth]{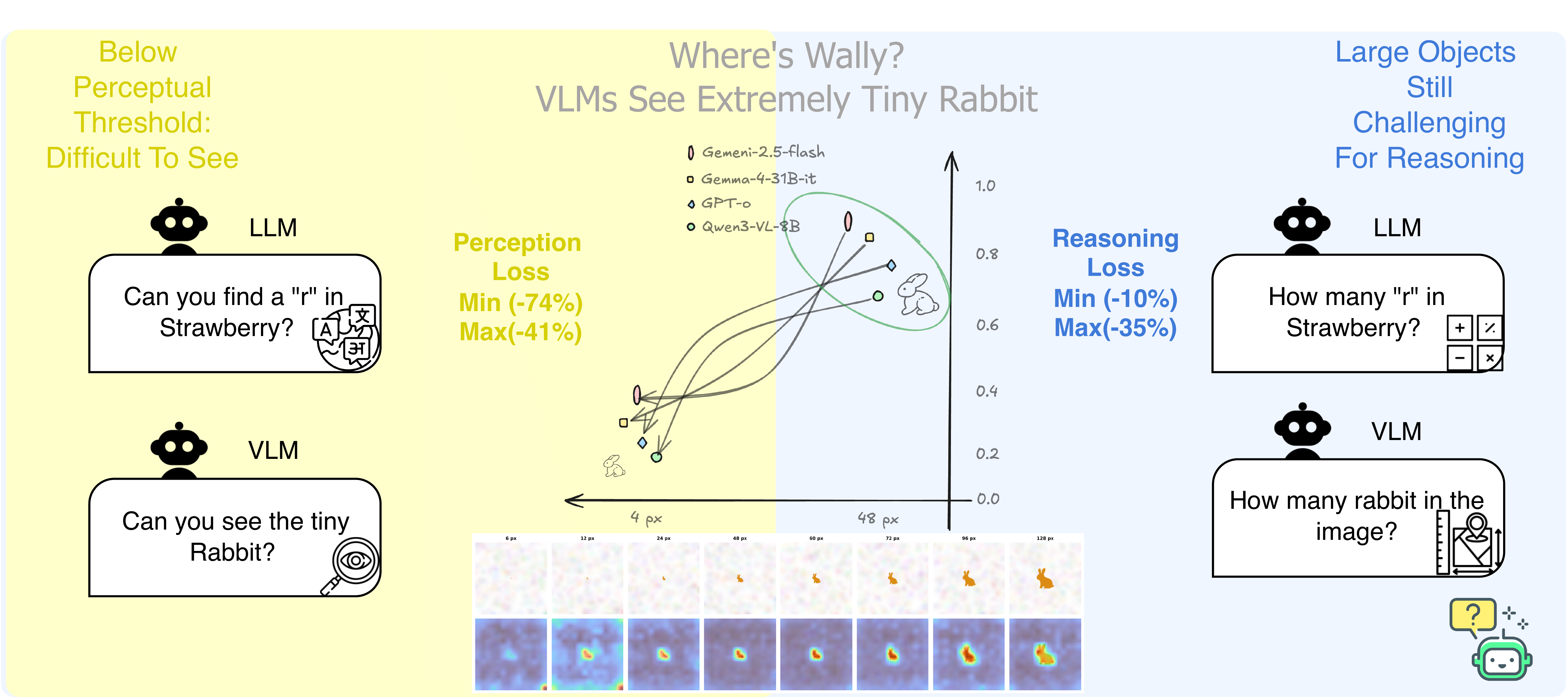}
  \caption{In this work, we evaluate VLMs from two complementary perspectives: visual recognition ability and visually grounded reasoning. Our results demonstrate that VLMs perform substantially worse on objects at \(4\) pixels than on those at \(48\) pixels, and that current models further struggle with complex spatial reasoning tasks, such as object counting. The bottom figure illustrates the degradation in attention response when the same rabbit stimulus is presented at a reduced scale.}
  \label{fig:finsightbench++}
\end{figure}

\section{Background and Related Work}

\paragraph{Input Image Resolution.} Historically, VLMs have evolved from low-resolution, fixed-size image inputs to higher-resolution and more flexible visual representations. Early VLMs typically operated on images of 224 or 336 pixels per side, a design that simplified implementation and controlled computational cost but limited performance on fine-grained tasks such as document understanding, and chart interpretation\cite{niu2025nativevisualunderstandingresolving,bordes2024introductionvisionlanguagemodeling}. More recent models aim to better preserve the original aspect ratio and visual content of input images by dynamic resizing and image tiling\cite{dehghani2023patchnpacknavit,wang2024qwen2vlenhancingvisionlanguagemodels}. Broadly, existing VLMs can be categorized into two groups. The first group consists of fixed-resolution models, where all input images are resized to a predefined resolution before being processed by the visual encoder. Examples include \textit{Kosmos-2}~\cite{kosmos-2,kosmos-1,metalm} at \(224 \times 224\), LLaVA-1.5~\cite{liu2023llava} at \(336 \times 336\), DeepSeek-VL2~\cite{lu2024deepseekvl,wu2024deepseekvl2mixtureofexpertsvisionlanguagemodels} at \(384 \times 384\), and Qwen-VL~\cite{bai2023qwenvlversatilevisionlanguagemodel} at \(448 \times 448\). The second group consists of dynamic-resolution models, which adapt their processing based on the original image size, aspect ratio, or a predefined pixel budget. For instance, Qwen2.5-VL~\cite{bai2025qwen25vltechnicalreport} constrains visual input using \texttt{min\_pixels} and \texttt{max\_pixels}, while InternVL2~\cite{chen2025expandingperformanceboundariesopensource} adopts a dynamic tiling strategy based on \(448 \times 448\) image patches. In practice, mainstream open-source VLMs typically adopt input resolutions ranging from 224 to 448 pixels as an empirical observation, which is also the reason why we chose 448 as the canvas size for our dataset. Closed-source models, by contrast, do not disclose a canonical internal image resolution; instead, they expose only API-level constraints---such as image file size, number of images or visual token budgets.

\paragraph{Large Unlimited, Small Limited to 1 Pixel.} Most VLMs' encoder typically relies on patch embedding: the input is divided into fixed-size patches, each patch is embedded into a visual token, and the resulting patch sequence is fed into the encoder together with positional information and padding masks. This design allows the model to process images of varying size with different patch size options. For example, Qwen-3VL \cite{qwen3technicalreport} and its SigLIP vision backbone \cite{10377550} use \(16 \times 16\) image patches. Other recent VLMs follow similar designs, including Llama-4-Scout-17B-16E-Instruct with \(14 \times 14\) patches, InternVL3.5-1B-Flash with \(12 \times 12\) patches, and Gemma-4 with \(16 \times 16\) patches. VLMs are effectively unconstrained for large objects due to cross-patch integration, but for small objects, recognition is bounded by the finest signal within a patch, approaching a lower limit of one pixel. From a user perspective, it is also important to understand the minimum pixel scale at which VLMs can reliably perceive objects.

\paragraph{Limited Benchmarking on Small-Object.} Contemporary visual question answering (VQA) evaluation has evolved beyond single-answer accuracy toward a more comprehensive assessment of fine-grained capabilities, including spatial relation understanding, optical character recognition (OCR), commonsense knowledge, logical reasoning, and visual mathematical reasoning. One main category of current VQA datasets evaluates general multimodal understanding~\cite{li2024seed2plus,li2023seed2,li2023seed}, examining whether a model can integrate images, text, and common sense knowledge to answer questions. MMBench~\cite{liu2024mmbenchmultimodalmodelallaround}, for instance, serves as a diagnostic benchmark that investigates various abilities even in mathematics. Existing studies further show that many mainstream VLMs are highly sensitive to object size in image question answering: smaller objects generally lead to lower answer accuracy, yet there is still no systematic quantitative analysis of VLMs' visual capabilities in the regime of extremely small objects~\cite{zhang2025mllmsknowlooktrainingfree}. Meanwhile, other work evaluates VLMs from the perspective of image resolution, where RC-Bench is proposed to examine model performance under different pixel densities, especially at extreme low resolutions, but it still does not directly quantify how small an object VLMs can reliably recognize or understand\cite{niu2025nativevisualunderstandingresolving}.

\paragraph{Seeing Does Not Guarantee Reasoning.}
Another important line of evaluation concerns specialized image-based reasoning, including exam-style question answering, solving mathematical problems with images, and interpretation of charts~\cite{lu2024mathvistaevaluatingmathematicalreasoning}. Other benchmarks target scenario-specific capabilities, such as WearVQA, a benchmark proposed by Meta, is designed to answer visual questions on wearable items and emphasizes 3D spatial reasoning~\cite{chang2025wearvqavisualquestionanswering}.
In addition, reliability-oriented benchmarks aim to diagnose hallucinations, assess robustness, and probe the limits of fine-grained capabilities~\cite{guan2024hallusionbenchadvanceddiagnosticsuite,li2023evaluatingobjecthallucinationlarge}.
In every benchmark, small objects always provide a particularly challenging setting when the target objects are too small compared to the background. A key question in image-text recognition remains open: how small an object a VLM can reliably perceive? Models may occasionally recognize small visual content, but fail in other cases; they may detect its presence without accurately localizing it. These inconsistencies point to a fundamental question regarding the minimum reliable resolving power of VLMs, and FineSightBench is designed to systematically investigate this limitation~\cite{lu2024mathvistaevaluatingmathematicalreasoning,Rahmanzadehgervi2024VisionLM}.

\section{From the Last Visible Pixel to Complex Reasoning}

\subsection{Motivation and Method Design}

When interacting with VLMs, users often wonder whether these systems can truly perceive all the information in an image, such as a medication brochure. This motivates our benchmark: although VLMs are often fluent in language-conditioned generation, they may remain limited in fine-grained visual perception in everyday use. Accordingly, our benchmark can probe the upper bound of VLM perceptual capability, namely the minimum visual scale at which they can still reliably recognize and reason about image content. We adopt the \textit{Test of Visual Perceptual Skills} (TVPS) as the foundation for design to evaluate human visual perceptual abilities~\cite{brown2018overview,martin2017tvps}. The TVPS aligns closely with several narrow abilities in Cattell--Horn--Carroll (CHC) theory, including closure flexibility and visual memory span~\cite{MCGREW20091}. It is also commonly used to support the assessment of visual perceptual deficits with learning difficulties or neurological disorders. 


Specifically, \textbf{Visual Discrimination} refers to the ability to detect subtle differences among visually similar shapes, symbols, or fine-grained details~\cite{Mikellidou2023}. In TVPS, this ability is typically assessed by asking participants to identify, from a set of visually similar candidates, the one that exactly matches a target image. \textbf{Spatial Relationships} concern the ability to judge the orientation, position, and relative arrangement of visual forms~\cite{Cheng2024SpatialRGPTGS,wang2024spatial,FRANCONERI2012210}. \textbf{Form Constancy} captures the ability to recognize the same figure despite changes in size, orientation, or surrounding context. \textbf{Sequential Memory} measures the ability to retain and reproduce a sequence of visual stimuli in the correct order. Finally, \textbf{Figure--Ground Perception} refers to the ability to identify a target figure embedded within a visually complex background~\cite{Dangelo2025}. 

Inspired by these concepts, we design the proposed benchmark to make it explicit which dimensions of visual perception are being tested. Rather than treating perception as a monolithic ability, our benchmark is structured around these established perceptual constructs from TVPS, so that each task probes a more specific aspect of fine-grained visual processing. Because contemporary VLMs are stateless~\cite{brown2020languagemodelsfewshotlearners} and therefore do not possess genuine memory in the human sense, the memory-related abilities in our benchmark are operationalized through single-image understanding combined with complex reasoning, rather than explicit multi-step memory retention~\cite{fan2025unveiling}.

\definecolor{groupbg}{gray}{0.90}

{
    \setlength{\textfloatsep}{5pt plus 1pt minus 2pt}
    \setlength{\dbltextfloatsep}{5pt plus 1pt minus 2pt}
    \begin{figure*}[htbp]
      \centering
    
      \small
      \setlength{\tabcolsep}{7pt}
      \renewcommand{\arraystretch}{1.22}
      \begin{adjustbox}{max width=\textwidth}
        \begin{tabular}{
          >{\raggedright\arraybackslash}p{4.1cm}
          >{\centering\arraybackslash}p{1.0cm}
          >{\raggedright\arraybackslash}p{8.3cm}
          >{\centering\arraybackslash}p{1.0cm}
          >{\centering\arraybackslash}p{1.4cm}
        }
          \toprule
          \textbf{Task Family} & \textbf{Abbr.} & \textbf{Representative Task Description} & \textbf{\# Obj.} & \textbf{\# Samples} \\
          \midrule
    
          \rowcolor{groupbg}
          \multicolumn{5}{l}{\textbf{FineSightBench-Perception} \; (\(7\) pixel levels: \(4, 8, 12, 16, 24, 32, 48\) px)} \\
          \addlinespace[2pt]
    
          \quad Letter Recognition
          & LTR
          & Identify the letter shown in the image.
          & 1
          & 700 \\
    
          \quad Animal Recognition
          & ANM
          & Identify the animal shown in the image.
          & 1
          & 700 \\
    
          \quad Block Recognition
          & BLK
          & Identify the block shown in the image.
          & 1
          & 700 \\
    
          \quad Color Block Recognition
          & CBL
          & Identify the block shown in the image and its color.
          & 1
          & 700 \\
    
          \quad Shape Recognition
          & SHP
          & Identify the geometric shape shown in the image.
          & 1
          & 700 \\
    
          \quad Text Recognition
          & TXT
          & What single English word is written in the image?
          & 1
          & 700 \\
    
          \midrule
    
          \rowcolor{groupbg}
          \multicolumn{5}{l}{\textbf{FineSightBench-Reasoning} \; (\(7\) pixel levels: \(4, 8, 12, 16, 24, 32, 48\) px)} \\
          \addlinespace[2pt]
    
          \quad Chain Reasoning
          & CHR
          & List the objects and their colors according to spatial order.
          & 2--10
          & 560 \\
    
          \quad Comparison Chain Reasoning
          & CMP
          & List the objects and their colors according to relative size.
          & 2--10
          & 560 \\
    
          \quad Counting Chain Reasoning
          & CNT
          & Report the number in each category and the total number.
          & 2--20
          & 840 \\
    
          \quad Blur Chain Reasoning
          & BLR
          & Report object counts in each category under blurred-background.
          & 2--10
          & 560 \\
    
          \quad Text Reading
          & TRD
          & List all the overlaid words from left to right in the Image.
          & 2--10
          & 560 \\
    
          \quad Text Counting
          & TCT
          & Total word count and count words containing a queried letter.
          & 2--10
          & 560 \\
    
          \bottomrule
        \end{tabular}
      \end{adjustbox}
    
      \vspace{0.2em}
    
      \includegraphics[width=\linewidth]{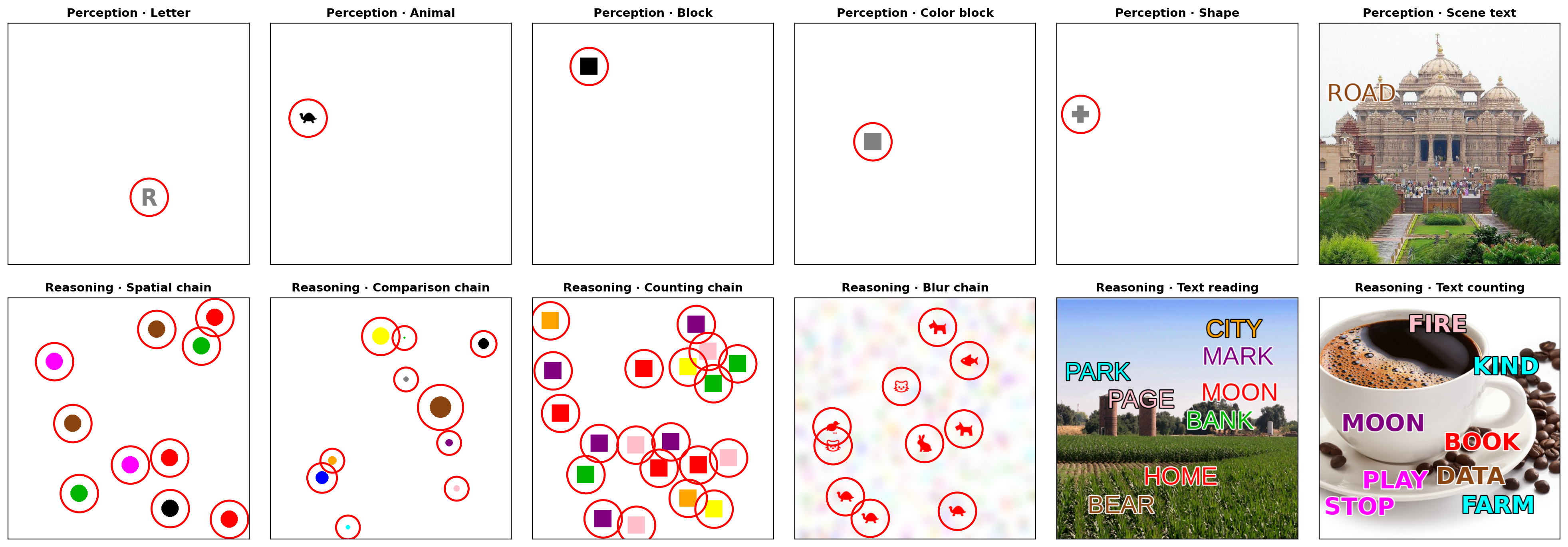}
    
      \caption{Overview of FineSightBench. Top: dataset statistics across task families; Bottom: representative task examples. Any visual highlighting of red circles is for illustration only and is not included in the images provided to the VLM.}
      \label{figtab:finesightbench_overview}
    \end{figure*}
    
}

\subsection{Dataset Construction}

\paragraph{Dataset Overview} We introduce a two-stage benchmark, termed \textbf{FineSightBench}, to evaluate the fine-grained visual perception and reasoning capabilities of VLMs under extreme visual conditions. Specifically, the benchmark examines whether VLMs can reliably perceive very small or visually weak targets, and whether they can further reason correctly based on such perceptual evidence. To this end, we construct two complementary datasets targeting two levels of competence: direct visual recognition and perception-conditioned reasoning.

\paragraph{Task Definition} As shown in Figure \ref{figtab:finesightbench_overview}, FineSightBench is a benchmark for evaluating fine-grained visual abilities under small-target conditions, based on the TVPS-inspired view that visual competence consists of multiple perceptual subabilities. It contains two hierarchical components, \textbf{FineSightBench-Perception} and \textbf{FineSightBench-Reasoning}, with 12 subtasks in total. The first evaluates basic perceptual abilities under extremely small-target conditions, including object detection, category discrimination, color identification, and text recognition, on targets such as letters, small animals, blocks, colored blocks, geometric shapes, and text across multiple pixel scales. The second evaluates perception-conditioned reasoning on multiple small objects, including tasks such as size comparison, spatial reasoning, counting, and reasoning under blur-based visual interference.

\paragraph{Dataset Construction}At the data-construction level, FineSightBench integrates two generation paradigms: synthetic data and TextWild data\cite{gupta2016syntheticdatatextlocalisation}. This is done to jointly satisfy the requirements of experimental controllability and diversity. The former typically renders visual objects such as letters, animals, geometric shapes, blocks, and dots on a $448 \times 448$ canvas with a white background or a controlled textured background, allowing strict control over variables including size, color, category, and spatial position, and making it well suited to isolate the effects of different factors on model performance. The latter overlays English words onto natural-scene backgrounds to simulate the task setting of recognizing tiny text in real-world images, thereby introducing practically relevant sources of difficulty such as background texture, edge interference, color contrast, and layout complexity. Meanwhile, the two data types follow a consistent set of generation principles. Objects are placed in a largely non-overlapping manner so that occlusion does not become the dominant source of difficulty; metadata such as size, position, color, bbox, and num\_targets are fully retained to support rigorous evaluation and error attribution; prompts explicitly constrain the candidate answer space to reduce the linguistic noise introduced by open-ended generation; and VLMs are uniformly required to produce outputs in JSON format so that perceptual failures, reasoning failures, and formatting failures can be distinguished during evaluation.

\subsection{Evaluation Protocol}

For all model evaluations, we adopt greedy decoding and require every response to strictly follow a predefined JSON schema. We also conduct an ablation study on different decoding settings in a later section for reproducibility with standardizes outputs for automatic parsing and scoring, and eliminates sampling variance so that differences more faithfully reflect visual understanding. We evaluate each model using four complementary metrics: Strict Accuracy (Exact), AUC-size, RT\(_X\), and Hallucination Rate. Together, they measure exact task performance, robustness across target scales, perceptual resolution threshold of a VLMs, and output reliability.

\paragraph{Strict Accuracy (Exact$\uparrow$).} For each sample \(i\), let \(y^{(i)}=\{y^{(i)}_k\}_{k=1}^{K_i}\) and \(\hat{y}^{(i)}=\{\hat{y}^{(i)}_k\}_{k=1}^{K_i}\) denote the reference and predicted JSON fields, where \(K_i\) is the number of evaluated fields. A prediction is considered correct if and only if all fields match exactly, and the task-level strict accuracy is defined as
\begin{equation}
\mathrm{Strict}_i=\mathbf{1}\!\left[\bigwedge_{k=1}^{K_i}\left(\hat{y}^{(i)}_k=y^{(i)}_k\right)\right],
\qquad
\mathrm{Acc}=\frac{1}{N}\sum_{i=1}^{N}\mathrm{Strict}_i.
\end{equation}

\paragraph{AUC-size (\(\uparrow\)).}
Let \(S=\{s_1,\dots,s_M\}\) denote the ordered set of evaluated target sizes, and let \(\mathrm{acc}(s_j)\in[0,1]\) be the strict accuracy at size \(s_j\). We define AUC-size as the normalized trapezoidal area under the accuracy--size curve:
\begin{equation}
\mathrm{AUC\text{-}size}
=
\frac{1}{s_{\max}-s_{\min}}
\sum_{j=1}^{M-1}
\frac{\mathrm{Acc}(s_j)+\mathrm{Acc}(s_{j+1})}{2}
\left(s_{j+1}-s_j\right),
\end{equation}
where \(s_{\min}=s_1\) and \(s_{\max}=s_M\). By construction, \(\mathrm{AUC\text{-}size}\in[0,1]\), and larger values indicate better accuracy maintained over a wider range of target sizes.

\paragraph{RT\(_X\) (\(\downarrow\)).}
Given a target accuracy threshold \(X\in(0,100)\), we define \(RT_X\) as the smallest target size at which the accuracy reaches \(X\)\%:
\begin{equation}
\mathrm{RT}_{X}
=
\min \left\{
s_1+\frac{X-\mathrm{Acc}(s_j)}{\mathrm{Acc}(s_{j+1})-\mathrm{Acc}(s_j)}(s_{j+1}-s_j)
\;\middle|\;
\mathrm{Acc}(s_j)<\frac{X}{100}\le \mathrm{Acc}(s_{j+1})
\right\},
\end{equation}
with \(\mathrm{RT}_X=s_1\) if \(\mathrm{Acc}(s_1)\ge X\), and \(\mathrm{RT}_X=s_{\max}+1\) if \(X\) is never reached. Smaller values indicate better fine-grained visual sensitivity.

\paragraph{Hallucination Rate (H$\downarrow$).}
In this paper, we define a hallucination as a \textbf{response that is not parseable as valid JSON and yields no extractable target field}. In particular, \( \mathrm{H@4px} \) denotes the hallucination rate at the smallest evaluated target scale. A higher value indicates that under extremely fine-grained visual conditions, the model fails not only semantically but also structurally, producing no valid structured output.

For overall model ranking, we primarily sort models by the mean AUC-size across tasks, while also reporting mean RT50, overall Strict Accuracy, and H@4px. This yields a compact evaluation protocol that jointly reflects perceptual robustness (AUC), perceptual threshold (RT50), task correctness (Strict Accuracy), and output reliability for smallest object(H@4px).

\section{Results}

\subsection{Average Performance}

\paragraph{Model-level} As shown in Table \ref{tab:finesightbench-leaderboard}, \textbf{Gemma-4-31B} is the overall top performer, achieving the highest AUC (\(78.8\)), the lowest RT50 (\(7.3\,\mathrm{px}\)), and either the best or second-best performance on multiple tasks. Its main weakness is hallucination at very small scales, with \(\mathrm{Hall}@4\mathrm{px}=13.4\%\), indicating a relatively high tendency to hallucinate when objects occupy only a few pixels. \textbf{Gemini-2.5-Flash} follows closely behind, obtaining the highest Mean\(\uparrow\) (\(74.8\)) and an almost saturated TCT score (Good at text counting). However, its RT50 is \(1.8\times\) that of Gemma-4-31B, meaning that it requires larger targets to reach the same \(50\%\) accuracy level. Overall, closed-source models still exhibit stronger aggregate capability. \textbf{GPT-4o} is distinguished by its extremely low hallucination rate, with \(\mathrm{Hall}@4\mathrm{px}=0.3\%\), indicating that it almost never fabricates answers. However, it trails the top-performing models by \(5\)--\(8\) points on reasoning-intensive tasks such as CHR\faHashtag and CMP\faBalanceScale.

%

\definecolor{bestcell}{HTML}{D4EDDA}
\definecolor{secondcell}{HTML}{FFF3CD}

\newcommand{\dataseticon}[1]{\makecell[c]{\large #1}}

\newcommand{\includelogo}[2][]{%
  \IfEndWith{#2}{.svg}{\includesvg[#1]{#2}}{\includegraphics[#1]{#2}}%
}

\newcommand{\modellogo}[2]{%
  \makecell[l]{%
    \IfFileExists{#1}{\raisebox{-0.15\height}{\includelogo[height=1.7ex]{#1}}\hspace{0.45em}}{}%
    #2%
  }%
}

\begin{table*}[!htbp]
\centering
\small
\renewcommand{\arraystretch}{1.40}
\setlength{\tabcolsep}{5pt}
\caption{
FineSightBench leaderboard. Per-task strict accuracy (\%), mean AUC over pixel sizes, mean $RT
_{50}$(px, lower is better), and JSON hallucination rate (\%). 
Bold on green = best per column; underline on yellow = second best.
Icons denote datasets/tasks: 
\faFont\ = LTR, 
\faShapes\ = SHP, 
\faPaw\ = ANM, 
\faThLarge\ = BLK, 
\faLink\ = CBL, 
\faFile\ = TXT, 
\faHashtag\ = CHR, 
\faBalanceScale\ = CMP, 
\faSortNumericDown\ = CNT, 
\faTint\ = BLR, 
\faRulerCombined\ = TRD, 
\faObjectGroup\ = TCT.
Company logos are shown before each model name.
}
\label{tab:finesightbench-leaderboard}

\begin{adjustbox}{max width=\textwidth}
\begin{tabular}{>{\raggedright\arraybackslash}p{3.8cm}ccccccccccccccccc}
\toprule
\multirow{2}{*}{\textbf{Model}} 
& \multicolumn{6}{c}{\textbf{FineSightBench Perception}} 
& \multicolumn{6}{c}{\textbf{FineSightBench Reasoning}} 
& \multicolumn{5}{c}{\textbf{Metrics}} \\
\cmidrule(lr){2-7}\cmidrule(lr){8-13}\cmidrule(lr){14-18}
& \dataseticon{\faFont}
& \dataseticon{\faShapes}
& \dataseticon{\faPaw}
& \dataseticon{\faThLarge}
& \dataseticon{\faLink}
& \dataseticon{\faFile} 
& \dataseticon{\faHashtag}
& \dataseticon{\faBalanceScale}
& \dataseticon{\faSortNumericDown}
& \dataseticon{\faTint}
& \dataseticon{\faRulerCombined}
& \dataseticon{\faObjectGroup}
& AUC$\uparrow$
& RT50$\downarrow$
& Hall.$\downarrow$
& Mean$\uparrow$
& H@4px$\downarrow$ \\
\midrule

\modellogo{photos/google.svg}{Gemma-4-31B} 
& \cellcolor{bestcell}\textbf{93.9} & \cellcolor{secondcell}\underline{86.9} & 69.8 & \cellcolor{bestcell}\textbf{100.0} & \cellcolor{bestcell}\textbf{100.0} & \cellcolor{bestcell}\textbf{89.9} & \cellcolor{bestcell}\textbf{46.8} & \cellcolor{bestcell}\textbf{42.1} & \cellcolor{secondcell}\underline{52.0} & \cellcolor{secondcell}\underline{83.0} & \cellcolor{bestcell}\textbf{52.3} & \cellcolor{secondcell}\underline{73.0} & \cellcolor{bestcell}\textbf{78.8} & \cellcolor{bestcell}\textbf{7.3} & 3.4 & \cellcolor{secondcell}\underline{74.5} & 13.4 \\

\modellogo{photos/google.svg}{Gemini-2.5-Flash} 
& \cellcolor{bestcell}\textbf{93.9} & 86.7 & \cellcolor{bestcell}\textbf{80.6} & 99.7 & 99.6 & 89.4 & 37.1 & 32.3 & 51.1 & 80.7 & \cellcolor{secondcell}\underline{47.0} & \cellcolor{bestcell}\textbf{89.3} & \cellcolor{secondcell}\underline{78.6} & 13.1 & 3.3 & \cellcolor{bestcell}\textbf{74.8} & 6.7 \\

\modellogo{photos/google.svg}{Gemma-4-26B-A4B} 
& 93.4 & 85.1 & 69.9 & \cellcolor{bestcell}\textbf{100.0} & 99.6 & \cellcolor{secondcell}\underline{89.7} & \cellcolor{secondcell}\underline{38.6} & 33.0 & \cellcolor{bestcell}\textbf{55.1} & \cellcolor{bestcell}\textbf{83.4} & 43.7 & 57.0 & 75.8 & \cellcolor{secondcell}\underline{11.1} & 2.1 & 71.4 & 9.1 \\

\modellogo{photos/openai.svg}{GPT-4o} 
& 93.0 & 81.6 & 67.4 & 99.9 & \cellcolor{secondcell}\underline{99.7} & 89.1 & 34.1 & \cellcolor{secondcell}\underline{34.5} & 48.7 & 77.9 & 44.4 & 64.4 & 74.4 & 14.0 & 0.0 & 70.3 & 0.3 \\

\modellogo{photos/Z.png}{GLM-4.6V-Flash} 
& 93.4 & 81.0 & 73.0 & 84.9 & 95.3 & 85.1 & 24.1 & 20.7 & 51.4 & 81.6 & 39.3 & 60.9 & 72.2 & 14.8 & 2.9 & 66.9 & 4.0 \\

\modellogo{photos/internvl.jpeg}{InternVL3.5-38B} 
& 87.9 & 72.6 & 64.7 & 85.6 & 99.3 & 82.7 & 26.1 & 18.8 & 38.3 & 69.6 & 39.1 & 56.3 & 68.5 & 16.4 & 0.0 & 62.6 & \cellcolor{bestcell}\textbf{0.0} \\

\modellogo{photos/qwen.svg}{Qwen3-VL-4B} 
& 93.6 & \cellcolor{bestcell}\textbf{87.0} & \cellcolor{secondcell}\underline{80.1} & 86.1 & 99.1 & 85.6 & 17.9 & 0.2 & 30.7 & 76.1 & 36.0 & 43.0 & 65.8 & 21.3 & 4.7 & 62.3 & 5.0 \\

\modellogo{photos/qwen.svg}{Qwen3-VL-8B} 
& 92.1 & 83.7 & 57.1 & 86.7 & \cellcolor{secondcell}\underline{99.7} & 86.9 & 7.3 & 4.8 & 34.6 & 78.8 & 37.7 & 48.3 & 64.8 & 19.1 & 5.8 & 60.9 & 7.1 \\

\modellogo{photos/internvl.jpeg}{InternVL3.5-30B-A3B} 
& 85.6 & 68.1 & 51.1 & 83.1 & 91.0 & 82.3 & 16.6 & 14.5 & 33.6 & 71.4 & 35.7 & 55.4 & 64.0 & 17.7 & \cellcolor{bestcell}\textbf{0.0} & 58.2 & \cellcolor{bestcell}\textbf{0.0} \\

\modellogo{photos/internvl.jpeg}{InternVL3.5-8B} 
& 86.1 & 64.1 & 48.9 & 76.9 & 98.1 & 82.6 & 16.4 & 15.9 & 31.2 & 68.2 & 37.4 & 48.3 & 63.1 & 18.9 & 0.0 & 56.9 & \cellcolor{bestcell}\textbf{0.0} \\

\modellogo{photos/internvl.jpeg}{InternVL3.5-14B} 
& 85.3 & 69.7 & 49.4 & 78.9 & 91.0 & 81.9 & 18.0 & 3.9 & 32.7 & 66.1 & 34.4 & 50.7 & 61.7 & 19.2 & 0.0 & 56.1 & \cellcolor{bestcell}\textbf{0.0} \\

\modellogo{photos/qwen.svg}{Qwen3-VL-30B-A3B} 
& 86.9 & 71.4 & 43.7 & 88.9 & 98.7 & 87.0 & 13.0 & 5.5 & 25.0 & 76.8 & 39.7 & 46.6 & 61.2 & 17.8 & 8.2 & 57.7 & 20.9 \\

\modellogo{photos/google.svg}{Gemma-4-E4B} 
& 92.6 & 78.3 & 62.9 & 93.0 & 97.4 & 87.3 & 20.2 & 16.6 & 30.4 & 53.0 & 34.3 & 6.1 & 61.0 & 24.3 & \cellcolor{bestcell}\textbf{0.0} & 56.9 & \cellcolor{bestcell}\textbf{0.0} \\

\modellogo{photos/internvl.jpeg}{InternVL3.5-2B} 
& 84.3 & 68.9 & 53.3 & 75.0 & 97.6 & 83.1 & 15.5 & 6.1 & 25.8 & 67.1 & 32.3 & 39.3 & 60.1 & 22.0 & \cellcolor{bestcell}\textbf{0.0} & 54.8 & \cellcolor{bestcell}\textbf{0.0} \\

\modellogo{photos/qwen.svg}{Qwen3-VL-2B} 
& 90.6 & 84.9 & 72.4 & 41.6 & 92.0 & 86.0 & 16.4 & 13.0 & 20.0 & 68.8 & 33.7 & 37.9 & 60.0 & 24.8 & \cellcolor{bestcell}\textbf{0.0} & 55.3 & \cellcolor{bestcell}\textbf{0.0} \\

\modellogo{photos/internvl.jpeg}{InternVL3.5-4B} 
& 86.1 & 65.7 & 50.1 & 73.0 & 93.7 & 82.0 & 11.8 & 9.8 & 28.7 & 70.4 & 33.6 & 32.0 & 59.5 & 23.8 & 0.0 & 53.8 & \cellcolor{bestcell}\textbf{0.0} \\

\modellogo{photos/google.svg}{Gemma-4-E2B} 
& 91.1 & 76.6 & 60.6 & 86.9 & 96.3 & 87.1 & 18.9 & 1.2 & 28.6 & 57.7 & 34.7 & 12.1 & 59.2 & 22.9 & \cellcolor{bestcell}\textbf{0.0} & 55.3 & \cellcolor{bestcell}\textbf{0.0} \\

\modellogo{photos/internvl.jpeg}{InternVL3.5-1B} 
& 85.3 & 60.3 & 44.1 & 77.1 & 98.6 & 82.4 & 0.0 & 0.2 & 20.0 & 56.4 & 28.6 & 46.0 & 55.0 & 22.8 & \cellcolor{bestcell}\textbf{0.0} & 51.0 & \cellcolor{bestcell}\textbf{0.0} \\

\bottomrule
\end{tabular}
\end{adjustbox}
\end{table*}

\begin{figure*}[h]
  \centering
  \hspace*{-0.05\textwidth}
  \includegraphics[width=0.95\textwidth]{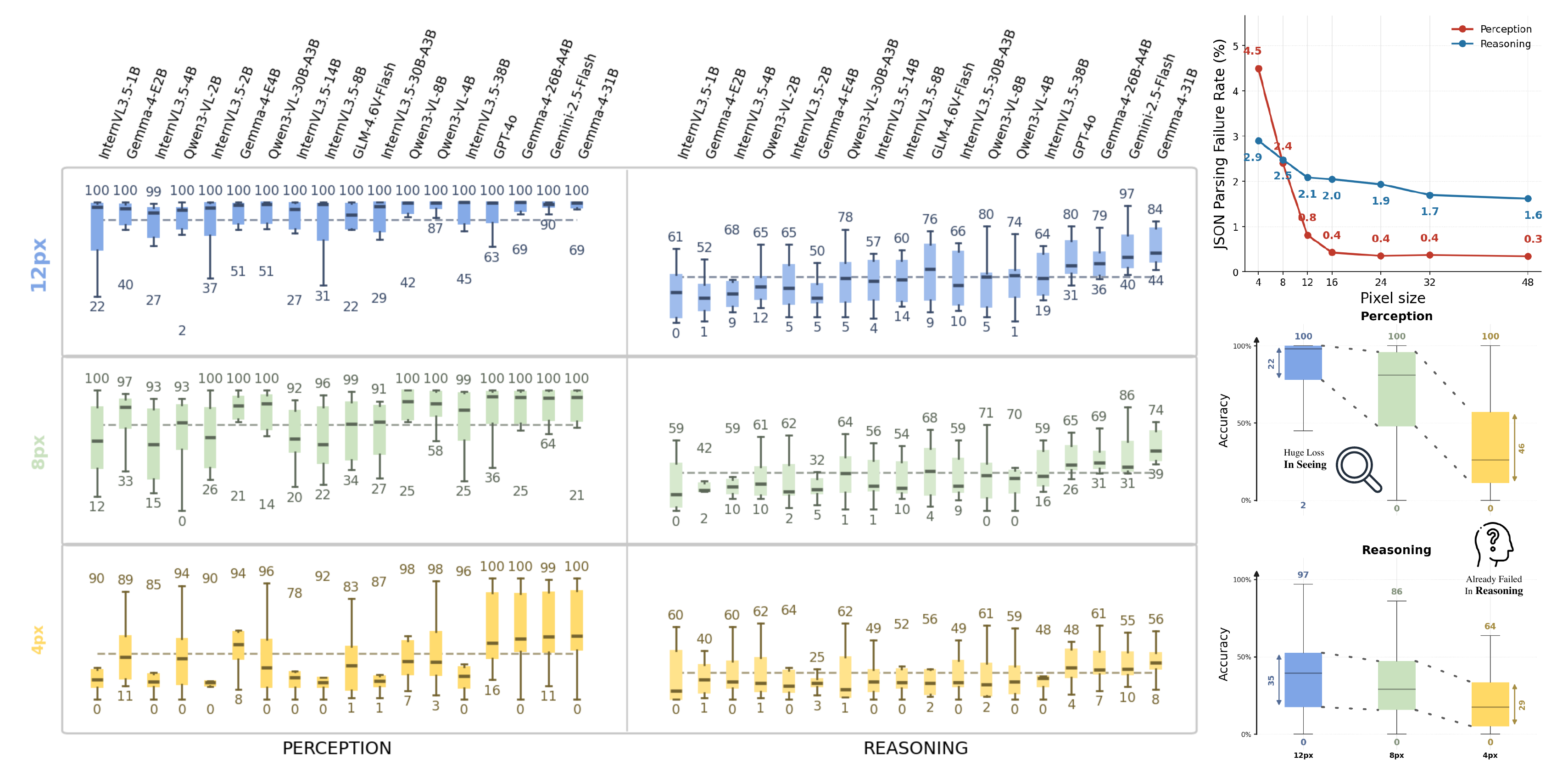}
  \caption{\textbf{Left:} Per-model strict-accuracy distributions at 4, 8, and 12,px, ordered by mean AUC-size. \textbf{Top right:} Mean JSON hallucination rate vs. pixel size. \textbf{Bottom right:} Accuracy distributions at 12, 8, and 4,px, showing clear degradation and increased variance as size decreases.}
  \label{fig:main-pixel-level-learderboard}
\end{figure*}

The \textbf{InternVL3.5} family and \textbf{Qwen3-VL-2B} can be regarded as the ``zero-hallucination'' group: all six InternVL3.5 variants, together with Gemma-4-E2B, Gemma-4-E4B, and Qwen3-VL-2B, achieve \(0\%\) on \(\mathrm{Hall}@4\mathrm{px}\). The trade-off is substantially weaker performance on reasoning-oriented tasks, especially CHR\faHashtag, CMP\faBalanceScale, TRD\faRulerCombined, and TCT\faObjectGroup, suggesting limited spatial reasoning ability. This suggests that, under structured output constraints, VLMs may optimize for format compliance rather than for reasoning toward the correct answer. The Qwen3-VL series shows an inverse scaling trend, with the smaller models outperforming the larger one. The larger model also exhibits the most severe hallucination under small-object conditions. Our later detailed analysis suggests that this behavior is largely due to the \(30\mathrm{B}\)-A3B model tending to refuse answering when it is uncertain. Within the InternVL3.5 family, performance is broadly monotonic with parameter scale, following \(38\mathrm{B} > 30\mathrm{B} > 14\mathrm{B} \approx 8\mathrm{B} > 4\mathrm{B} > 2\mathrm{B} > 1\mathrm{B}\), which constitutes the most stable scaling trend.
\paragraph{Task-level} Perception-oriented tasks are largely saturated: for tasks that primarily require ``seeing,'' the leading models generally achieve accuracies above \(90\%\). In contrast, reasoning is the true dividing line. For CHR\faHashtag and CMP\faBalanceScale, the best scores across the entire benchmark are only \(46.8\) and \(42.1\), respectively, while most models score below \(20\). This indicates that current models still perform poorly on spatial comparative reasoning problems that are relatively simple for humans. TCT\faObjectGroup is extremely polarized, with performance ranging from \(89.3\) for Gemini-2.5-Flash to only \(6.1\) for Gemma-4-E4B. Because it requires both OCR and counting ability, it is the most sensitive indicator for distinguishing models in terms of overall capability. The task difficulty follows a clear hierarchy: perception tasks (LTR \faFont / BLK \faThLarge / CBL \faLink) are near saturation; counting tasks (CNT \faSortNumericDown / BLR \faTint / TCT \faObjectGroup) show moderate difficulty; ordering and comparison tasks (CHR \faHashtag / CMP \faBalanceScale / TRD \faRulerCombined) remain highly challenging. This validates the design of FineSightBench. The gap between \(\mathrm{Hall}@4\,\mathrm{px}\) and AUC suggests that improving robustness at the \(4\,\mathrm{px}\) scale is the most effective direction for optimization.

\paragraph{Pixel-Level} As shown in Fig.~\ref{fig:main-pixel-level-learderboard}, in \(12\,\mathrm{px}\), perception tasks are already close to saturation, and the main differences between models are largely determined by whether they still exhibit weaknesses in the most challenging perception subtask. At \(8\,\mathrm{px}\), perceptual capability becomes the key stage at which model performance begins to diverge. By \(4\,\mathrm{px}\), only a few top-performing models can still maintain effective performance on some coarse-grained perception tasks, while most others fail almost completely. In contrast, reasoning remains a more prominent bottleneck throughout: even at \(12\,\mathrm{px}\), it is far from saturated, and substantial variation exists across different reasoning subtasks. Although some strong models retain limited reasoning ability at \(8\,\mathrm{px}\), performance at \(4\,\mathrm{px}\) drops to nearly the lower bound for almost all models. Therefore, the advantage of stronger models is primarily reflected in two aspects: they can still ``see'' at extremely small scales, and they can continue to reason once they are able to see. The hallucination rate, as shown in Fig.~\ref{fig:main-pixel-level-learderboard}, is higher for perception tasks. This is because the dataset contains only a single target of size 44~pixels, making VLMs more likely to abstain from answering. In contrast, reasoning tasks typically involve 2--10 targets, so VLMs tend to provide an answer after locating targets, regardless of correctness.
\subsection{Failure Mode Analysis}

\begin{wraptable}{r}{0.45\textwidth}
\vspace{-1.2\baselineskip} 
\small
\setlength{\tabcolsep}{5pt}
\caption{Top-5 @4px hallucination \% models.}
\label{tab:halluc-top5-combined-by-4px}
\centering
\resizebox{\linewidth}{!}{%
\begin{tabular}{llccccccc}
\toprule
Split & Model & 4\,px & 8\,px & 12\,px & 16\,px & 24\,px & 32\,px & 48\,px \\
\midrule
\multirow{5}{*}{Perception} & Qwen3-VL-30B-A3B & \cellcolor[HTML]{67000C}\textcolor{white}{\textbf{33.3}} & \cellcolor[HTML]{F5543C}\textcolor{white}{\textbf{18.5}} & \cellcolor[HTML]{FCB89D}8.7 & \cellcolor[HTML]{FCC9B4}6.7 & \cellcolor[HTML]{FCCDB9}6.3 & \cellcolor[HTML]{FCCCB7}6.5 & \cellcolor[HTML]{FDCEBA}6.2 \\
 & Gemma-4-31B & \cellcolor[HTML]{C7171C}\textcolor{white}{25.4} & \cellcolor[HTML]{FB7A5A}14.9 & \cellcolor[HTML]{FEE1D3}4.0 & \cellcolor[HTML]{FEF1EA}0.8 & \cellcolor[HTML]{FFF5F0}0.0 & \cellcolor[HTML]{FFF5F0}0.0 & \cellcolor[HTML]{FFF5F0}0.0 \\
 & Gemma-4-26B-A4B & \cellcolor[HTML]{F86144}\textcolor{white}{17.3} & \cellcolor[HTML]{FCB89D}8.7 & \cellcolor[HTML]{FEEDE4}1.7 & \cellcolor[HTML]{FEF4EF}0.2 & \cellcolor[HTML]{FFF5F0}0.0 & \cellcolor[HTML]{FFF5F0}0.0 & \cellcolor[HTML]{FFF5F0}0.0 \\
 & Gemini-2.5-Flash & \cellcolor[HTML]{FEE5DA}3.0 & \cellcolor[HTML]{FEF3ED}0.5 & \cellcolor[HTML]{FFF5F0}0.0 & \cellcolor[HTML]{FFF5F0}0.0 & \cellcolor[HTML]{FFF5F0}0.0 & \cellcolor[HTML]{FFF5F0}0.0 & \cellcolor[HTML]{FFF5F0}0.0 \\
 & GLM-4.6V-Flash & \cellcolor[HTML]{FEF1EB}0.7 & \cellcolor[HTML]{FFF5F0}0.0 & \cellcolor[HTML]{FFF5F0}0.0 & \cellcolor[HTML]{FFF5F0}0.0 & \cellcolor[HTML]{FFF5F0}0.0 & \cellcolor[HTML]{FFF5F0}0.0 & \cellcolor[HTML]{FFF5F0}0.0 \\
\midrule
\multirow{5}{*}{Reasoning} & Qwen3-VL-8B & \cellcolor[HTML]{FB7E5E}14.5 & \cellcolor[HTML]{FB8F6F}12.9 & \cellcolor[HTML]{FC9474}\textbf{12.3} & \cellcolor[HTML]{FB8F6F}\textbf{12.9} & \cellcolor[HTML]{FC9474}\textbf{12.3} & \cellcolor[HTML]{FCA284}\textbf{10.9} & \cellcolor[HTML]{FCB69C}8.7 \\
 & Gemini-2.5-Flash & \cellcolor[HTML]{FCA386}10.7 & \cellcolor[HTML]{FC9E80}11.3 & \cellcolor[HTML]{FCC2AB}7.5 & \cellcolor[HTML]{FDCFBC}6.1 & \cellcolor[HTML]{FEE0D2}4.1 & \cellcolor[HTML]{FEE3D7}3.4 & \cellcolor[HTML]{FEEDE4}1.6 \\
 & Qwen3-VL-4B & \cellcolor[HTML]{FCA78A}10.4 & \cellcolor[HTML]{FCB094}9.5 & \cellcolor[HTML]{FCB196}9.3 & \cellcolor[HTML]{FCB094}9.5 & \cellcolor[HTML]{FCA88B}10.2 & \cellcolor[HTML]{FCB499}9.1 & \cellcolor[HTML]{FCAB8E}\textbf{10.0} \\
 & GLM-4.6V-Flash & \cellcolor[HTML]{FCC2AB}7.5 & \cellcolor[HTML]{FCCAB6}6.6 & \cellcolor[HTML]{FDD0BD}5.9 & \cellcolor[HTML]{FCCDB9}6.2 & \cellcolor[HTML]{FDD5C3}5.4 & \cellcolor[HTML]{FDDDCE}4.5 & \cellcolor[HTML]{FDDACA}4.8 \\
 & Qwen3-VL-30B-A3B & \cellcolor[HTML]{FCC2AB}7.5 & \cellcolor[HTML]{FEE3D6}3.6 & \cellcolor[HTML]{FEE7DC}2.9 & \cellcolor[HTML]{FEE8DE}2.5 & \cellcolor[HTML]{FEE5DA}3.0 & \cellcolor[HTML]{FEE7DD}2.7 & \cellcolor[HTML]{FEE2D5}3.7 \\
\bottomrule
\end{tabular}%
}
\end{wraptable}

We first report the top 5 models ranked by hallucination rate at $4\,\mathrm{px}$, as shown in Table~\ref{tab:halluc-top5-combined-by-4px}.
The top and bottom sections of the table correspond to Perception and Reasoning, respectively,
with rows sorted in descending order within each split; \textbf{bold} denotes the highest
hallucination rate in each column across the two splits. In general, we observe that relatively
larger models tend to exhibit more severe hallucinations under extreme low-resolution conditions.

\noindent This finding is particularly pronounced for \texttt{Qwen3-VL-30B} despite its scale. To better
understand this behavior, we further perform a sample-level analysis. We found that, when the
target becomes extremely small, this model often refuses to answer rather than producing an
incorrect but well-formed prediction; we term this failure
mode \textbf{\emph{Refusal Failure}}, as shown in Figure \ref{fig:refusal_example}. Since other models are still able
to produce correct answers on the same samples $4\,\mathrm{px}$, we rule out the possibility
that the targets are inherently too small to be recognized and instead attribute this phenomenon
to a model-specific failure pattern near the lower bound of visual resolution. More broadly,
this finding suggests that VLM failure patterns in this regime are heterogeneous: some models
tend to hallucinate structured but incorrect outputs, whereas others collapse into refusal
behavior and fail to provide any task-compliant response.

\newlength{\fmTableCaptionVSpace}
\setlength{\fmTableCaptionVSpace}{6pt}

\begin{table}[htbp]
\centering
\small

\begin{minipage}[t]{0.69\linewidth}
\centering
\vspace{0pt}
\setlength{\tabcolsep}{4pt}
\resizebox{\linewidth}{!}{%
\begin{tabular}{lrrrrrrrr}
\toprule
 & 4px & 8px & 12px & 16px & 24px & 32px & 48px & MEAN \\
\midrule
Refusal Failure(REF) & \cellcolor[HTML]{FEDBCC} 2.11 & \cellcolor[HTML]{FEE8DD} 1.21 & \cellcolor[HTML]{FFF0E9} 0.42 & \cellcolor[HTML]{FFF3ED} 0.22 & \cellcolor[HTML]{FFF3ED} 0.18 & \cellcolor[HTML]{FFF3ED} 0.19 & \cellcolor[HTML]{FFF3ED} 0.19 & \cellcolor[HTML]{FFEEE6} 0.65 \\
Format Violation (FMT) & \cellcolor[HTML]{FCC1A8} 3.48 & \cellcolor[HTML]{FDCBB6} 2.95 & \cellcolor[HTML]{FDCEBB} 2.81 & \cellcolor[HTML]{FDD3C1} 2.56 & \cellcolor[HTML]{FDD7C6} 2.37 & \cellcolor[HTML]{FED9C9} 2.25 & \cellcolor[HTML]{FED9C9} 2.24 & \cellcolor[HTML]{FDD1BE} 2.67 \\
Misclassification (MSC) & \cellcolor[HTML]{67000D} \textcolor{white}{22.89} & \cellcolor[HTML]{CC191E} \textcolor{white}{11.14} & \cellcolor[HTML]{FC8161} 6.44 & \cellcolor[HTML]{FCA285} 4.90 & \cellcolor[HTML]{FEDCCD} 2.09 & \cellcolor[HTML]{FEE8DD} 1.21 & \cellcolor[HTML]{FFEFE8} 0.57 & \cellcolor[HTML]{FB7555} 7.03 \\
Color Confusion (CLR) & \cellcolor[HTML]{FFEEE7} 0.62 & \cellcolor[HTML]{FFF2EC} 0.28 & \cellcolor[HTML]{FFF3ED} 0.20 & \cellcolor[HTML]{FFF3ED} 0.23 & \cellcolor[HTML]{FFF4EE} 0.12 & \cellcolor[HTML]{FFF4EF} 0.11 & \cellcolor[HTML]{FFF3ED} 0.20 & \cellcolor[HTML]{FFF2EC} 0.25 \\
OCR Error (OCE) & \cellcolor[HTML]{FB7D5D} 6.59 & \cellcolor[HTML]{FEE1D4} 1.79 & \cellcolor[HTML]{FFF5F0} 0.03 & \cellcolor[HTML]{FFF5F0} 0.03 & \cellcolor[HTML]{FFF5F0} 0.00 & \cellcolor[HTML]{FFF5F0} 0.00 & \cellcolor[HTML]{FFF5F0} 0.00 & \cellcolor[HTML]{FEE8DD} 1.21 \\
Numeracy Error (NUM) & \cellcolor[HTML]{67000D} \textcolor{white}{16.78} & \cellcolor[HTML]{75030F} \textcolor{white}{14.56} & \cellcolor[HTML]{A60F15} \textcolor{white}{13.01} & \cellcolor[HTML]{A91016} \textcolor{white}{12.93} & \cellcolor[HTML]{C8171C} \textcolor{white}{11.42} & \cellcolor[HTML]{D72322} \textcolor{white}{10.65} & \cellcolor[HTML]{E22E27} \textcolor{white}{10.03} & \cellcolor[HTML]{AC1117} \textcolor{white}{12.77} \\
Sequence Error (SEQ) & \cellcolor[HTML]{67000D} \textcolor{white}{17.87} & \cellcolor[HTML]{67000D} \textcolor{white}{15.37} & \cellcolor[HTML]{900A12} \textcolor{white}{13.72} & \cellcolor[HTML]{AC1117} \textcolor{white}{12.77} & \cellcolor[HTML]{A50F15} \textcolor{white}{13.12} & \cellcolor[HTML]{A81016} \textcolor{white}{12.98} & \cellcolor[HTML]{7A0510} \textcolor{white}{14.38} & \cellcolor[HTML]{7C0510} \textcolor{white}{14.32} \\
\bottomrule
\end{tabular}%
}

\vspace{\fmTableCaptionVSpace}
\captionsetup{type=table}
\captionof{table}{Failure-mode rate (\%) at each pixel size (rows: failure mode; columns: target size in pixels; cell = \#failures Rate).}
\label{tab:fm_pixel}
\end{minipage}%
\hfill
\begin{minipage}[t]{0.29\linewidth}
\centering
\vspace{-6pt}
\includegraphics[width=\linewidth]{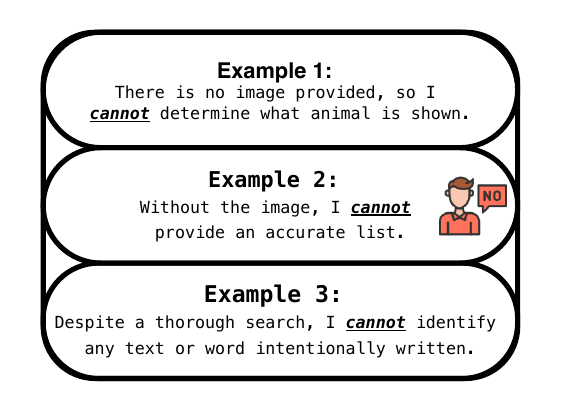}

\captionsetup{type=figure}
\captionof{figure}{Refusal examples shown in large VLMs.}
\label{fig:refusal_example}
\end{minipage}
\end{table}

\paragraph{Failure Vs Pixel Size} We also examine other failure modes, as summarized in Table~\ref{tab:fm_pixel}. \textbf{REF} denotes cases where the model directly refuses to answer and does not produce a usable JSON output. \textbf{FMT} refers to cases where the model attempts to answer but breaks the required format, resulting in an unparseable JSON response without explicit refusal expressions. As a reminder, the previously defined hallucination rate is calculated as the sum of REF and FMT. \textbf{MSC} indicates cases where the format is correct but the object category is misclassified. \textbf{CLR} refers to cases where the object is recognized correctly but its color is predicted incorrectly. \textbf{OCE} denotes OCR-related errors, where the model misreads text. \textbf{NUM} corresponds to counting errors, where the number of objects is predicted incorrectly. \textbf{SEQ} refers to errors in the list content or ordering. The results show a clear shift from perception-related failures to structural reasoning failures as target size increases from \(4\text{px}\) to \(48\text{px}\). For very small targets, recognition errors dominate, with Misclassification and OCR Error reaching \(22.89\%\) and \(6.59\%\), respectively, but both drop to near zero at larger pixel sizes. In contrast, Numeracy Error and Sequence Error remain consistently high across all settings, indicating that the model's main bottleneck in multi-object scenes lies in counting objects correctly and organizing them in the required order. More detailed definitions of failure modes are provided in the appendix.


\begin{wraptable}{r}{0.35\textwidth}
\vspace{-\intextsep}
  \centering
  \caption{Failure mode rates (\%) across different object counts.}
  \label{tab:failure_modes_vs_object_count}
  \begin{adjustbox}{width=\linewidth}
    \begin{tabular}{lrrrrrr}
      \toprule
       & 2 & 4 & 6 & 8 & 10 & MEAN \\
      \midrule
      REF & \cellcolor[HTML]{FFF5F0}0.02 & \cellcolor[HTML]{FFF5F0}0.02 & \cellcolor[HTML]{FFF5F0}0.00 & \cellcolor[HTML]{FFF5F0}0.01 & \cellcolor[HTML]{FFF5F0}0.04 & \cellcolor[HTML]{FFF5F0}0.02 \\
      FMT & \cellcolor[HTML]{FCB89E}3.87 & \cellcolor[HTML]{FCB398}4.11 & \cellcolor[HTML]{FFF0E8}0.51 & \cellcolor[HTML]{FC8969}6.05 & \cellcolor[HTML]{BD151A}\textcolor{white}{11.92} & \cellcolor[HTML]{FC997A}5.29 \\
      MSC & \cellcolor[HTML]{FFF5F0}0.00 & \cellcolor[HTML]{FFF5F0}0.00 & \cellcolor[HTML]{FFF5F0}0.00 & \cellcolor[HTML]{FFF5F0}0.00 & \cellcolor[HTML]{FFF5F0}0.00 & \cellcolor[HTML]{FFF5F0}0.00 \\
      CLR & \cellcolor[HTML]{FFF5F0}0.00 & \cellcolor[HTML]{FFF5F0}0.00 & \cellcolor[HTML]{FFF5F0}0.00 & \cellcolor[HTML]{FFF5F0}0.00 & \cellcolor[HTML]{FFF5F0}0.00 & \cellcolor[HTML]{FFF5F0}0.00 \\
      OCE & \cellcolor[HTML]{FFF5F0}0.00 & \cellcolor[HTML]{FFF5F0}0.00 & \cellcolor[HTML]{FFF5F0}0.00 & \cellcolor[HTML]{FFF5F0}0.00 & \cellcolor[HTML]{FFF5F0}0.00 & \cellcolor[HTML]{FFF5F0}0.00 \\
      NUM & \cellcolor[HTML]{D11E1F}\textcolor{white}{10.93} & \cellcolor[HTML]{67000D}\textcolor{white}{19.80} & \cellcolor[HTML]{67000D}\textcolor{white}{30.11} & \cellcolor[HTML]{67000D}\textcolor{white}{27.90} & \cellcolor[HTML]{67000D}\textcolor{white}{28.08} & \cellcolor[HTML]{67000D}\textcolor{white}{23.36} \\
      SEQ & \cellcolor[HTML]{67000D}\textcolor{white}{18.30} & \cellcolor[HTML]{67000D}\textcolor{white}{31.22} & \cellcolor[HTML]{67000D}\textcolor{white}{36.35} & \cellcolor[HTML]{67000D}\textcolor{white}{40.70} & \cellcolor[HTML]{67000D}\textcolor{white}{38.52} & \cellcolor[HTML]{67000D}\textcolor{white}{33.02} \\
      \bottomrule
    \end{tabular}
  \end{adjustbox}
\end{wraptable}

\paragraph{Failure Vs \#Objects} As the number of target objects increases from \(2\) to \(10\), the model's overall failure rate rises substantially for FMT, NUM and SEQ, as shown in Table \ref{tab:failure_modes_vs_object_count}, with errors concentrated mainly in two structural categories: Sequence Error and Numeracy Error. Sequence Error is the dominant failure mode throughout, increasing from \(18.3\%\) at \(2\) objects to \(40.7\%\) at \(8\) objects, nearly doubling. Numeracy Error also grows markedly, rising from \(10.9\%\) at \(2\) objects to \(30.1\%\) at \(6\) objects, and then remaining high at around \(28\%\)--\(30\%\) for \(6\) to \(10\) objects. By contrast, recognition-related errors, including misclassification, color confusion, OCR error, and refusal, remain close to zero across all settings, suggesting that the main challenge in multi-object tasks lies not in recognizing individual objects but in organizing them then doing logical reasoning. In addition, Format Violation increases sharply under heavy visual crowding, rebounding from \(0.51\%\) at \(6\) objects to \(11.92\%\) at \(10\) objects, which indicates that the model becomes more likely to deviate as scene complexity grows.

\subsection{Ablation Study}

In the previous setting, we used strict exact match and greedy decoding which may underestimate semantic performance. Two more ablation studies are conducted on the choice of decoding settings and metrics.

First, we examine whether model rankings are sensitive to the evaluation criterion by comparing \textbf{\texttt{exact}} match with two more permissive alternatives:\textbf{ \texttt{set-match}}, which ignores differences in field order and surface form but requires the same tokens with the same multiplicities, and \textbf{\texttt{overlap}}, which measures token overlap with the gold answer and penalizes both omissions and spurious content. Second, we examine whether the decoding strategy affects the relative performance of the model by evaluating each model in three settings: \texttt{greedy}, low-temperature sampling \((T=0.1)\), and high-temperature sampling \((T=1.0)\), following prior work showing that the sampling strategy can potentially influence the performance of LLM~\cite{LI2025242}.



\begin{figure*}[h]
  \centering
  \hspace{-0.5cm}
  \includegraphics[width=1.02\linewidth]{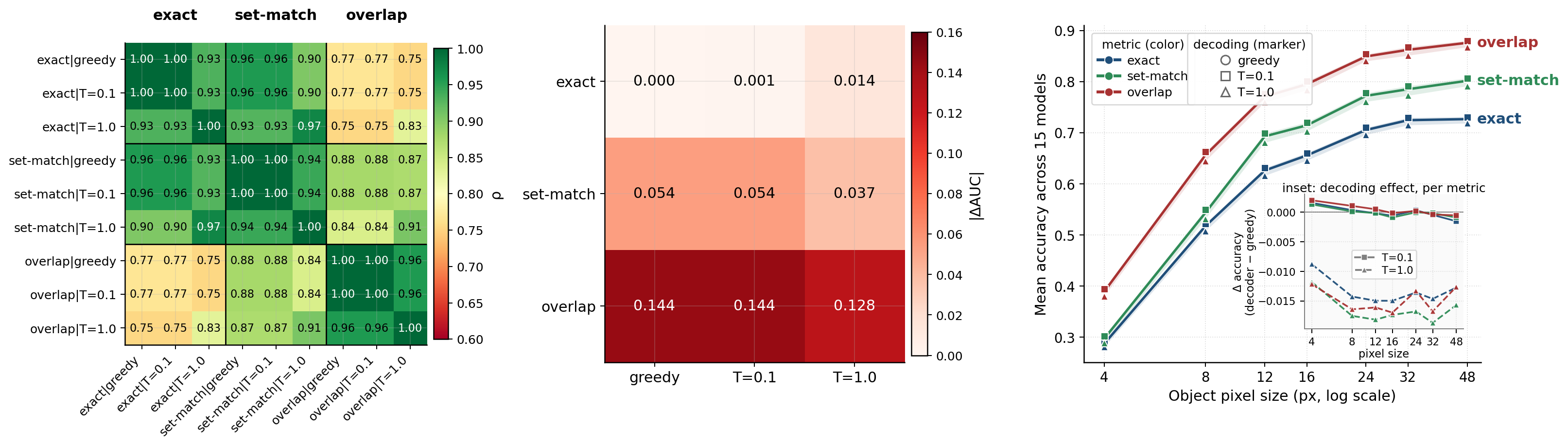}
  \caption{%
    \textbf{Joint ablation of evaluation metric and decoding strategy.}
    \textbf{Left:} Spearman $\rho$ defined in equation \ref{eq:spearman-correlation} across the 9
    (metric, decoding) schemes.
    \textbf{Middle:} mean $|\Delta\text{AUC}|$ defined in equation \ref{eq:delta_auc} vs.\ \emph{exact/greedy}:
    swapping decoder costs ${\le}0.015$.
    \textbf{Right:} accuracy vs.\ object pixel size; color = metric,
    marker = decoder. The three decoders nearly coincide while the
    three metrics stay cleanly separated; the inset shows decoder
    residuals.
  }
  \label{fig:ablation_joint}
\end{figure*}

Figure \ref{fig:ablation_joint} shows that the evaluation metric is the main source of variation, while the decoding strategy has a relatively minor effect. The metric causes only limited changes in model ranking and decoding does not materially affect ordering. The right panel further confirms that the ordering \texttt{overlap} \(>\) \texttt{set-match} \(>\) \texttt{exact} is preserved across the entire range from \(4\)px to \(48\)px, and the extremely thin bands, in which curves under different decoding strategies nearly collapse onto a single line, indicate that decoding remains locally stable. Among the three decoding settings, \(T=1.0\) is generally slightly below \texttt{greedy}, suggesting a consistent but modest degradation, while \(T=0.1\) is nearly indistinguishable from \texttt{greedy}. In general, these findings support the validity of the experimental design and indicate that sampling-based decoding has little influence on absolute performance or model ranking.

\section{Conclusion And Limitations}

This paper addresses an underexplored question: the effective visual resolution lower bound of contemporary VLMs. The experiments conducted reveal three key findings. First, a clear gap exists between perception and reasoning: basic perceptual performance saturates around 12 pixels, while reasoning remains far from saturation at all scales. This exposes a fundamental mismatch between seeing and reasoning, and highlights a limitation of benchmarks that mix the two concepts together. Second, inter-model differences stem more from pipeline stability than perceptual sensitivity: top models consistently maintain the full chain from fine-grained perception to structured reasoning at extremely small scales, rather than achieving sporadic recognition. Third, failure analysis shows that NumeracyError and SequenceError persist across all scales, indicating that multi-object reasoning and structured processing under low-resolution conditions remain major challenges for current VLMs.

Despite these contributions, this work has several limitations. First, although FineSightBench is designed to balance controllability, a considerable portion of the benchmark is still constructed from synthetic data, which inevitably leaves a distribution gap with complex real-world visual scenes. Second, our evaluation mainly focuses on target scale, structured output correctness, and end-task performance, while offering only limited insight into the internal mechanisms of attention. Third, we use unified prompting, greedy decoding, and strict JSON matching for reproducibility; however, this may underestimate performance under more permissive or alternative settings. Although our ablation results suggest that decoding strategy has only a minor effect on model ranking, the choice of evaluation metric can still affect absolute scores. Finally, this work focuses on fine-grained perception and reasoning in static images for current VLMs, and does not cover settings such as video, multi-image input, tool use, or adaptive visual processing. These factors may affect practical performance and remain important directions for future work.



\bibliographystyle{plainnat}
\bibliography{references}


\appendix

\newpage

\section{Supplementary Metrics For Ablation}

We investigate whether model rankings change under alternative correctness criteria while holding the decoding strategy fixed to \texttt{greedy}. Let \(\mathcal{M}=\{m_1,\dots,m_N\}\) denote the set of evaluated models, where \(N=18\), and let
\begin{equation}
\mathcal{C}=\{\texttt{\textbf{exact}},\ \texttt{set-match},\ \texttt{overlap}\}
\end{equation}
denote the set of correctness criteria. The correctness criterion is treated as the independent variable. For each model \(m_i\in\mathcal{M}\) and each criterion \(c\in\mathcal{C}\), we compute an AUC-size score \(A_{i,c}\in[0,1]\). This produces an evaluation matrix
\begin{equation}
A\in[0,1]^{N\times |\mathcal{C}|},
\end{equation}
which in our setting has dimension \(18\times 3\).

To quantify the extent to which model rankings depend on the correctness criterion, we take \(\texttt{exact}\) as the reference and compare each alternative criterion \(c\in\mathcal{C}\setminus\{\texttt{exact}\}\) against it along two dimensions. First, ranking consistency is measured by Spearman's rank correlation coefficient:
\begin{equation}
\rho_c=\mathrm{Spearman}\bigl((A_{i,\texttt{exact}})_{i=1}^N,\ (A_{i,c})_{i=1}^N\bigr).
\label{eq:spearman-correlation}
\end{equation}
Second, absolute score variation is measured by the per-model AUC difference
\begin{equation}
\Delta_{i,c} AUC=A_{i,c}-A_{i,\texttt{exact}},
\label{eq:delta_auc}
\end{equation}
together with its mean and maximum absolute values:
\begin{equation}
\mathrm{MeanAbsDiff}_c=\frac{1}{N}\sum_{i=1}^N |\Delta_{i,c}|,
\qquad
\mathrm{MaxAbsDiff}_c=\max_{1\leq i\leq N} |\Delta_{i,c}|.
\end{equation}

We now formalize the correctness criteria. For each evaluation instance, let \(G\) and \(P\) denote the gold output and the predicted output, respectively. After tokenization and normalization, both are represented as finite multisets over a token universe \(\mathcal{V}\). Equivalently, \(G\) and \(P\) may be viewed as multiplicity functions
\begin{equation}
G,P:\mathcal{V}\to\mathbb{N},
\end{equation}
where \(G(t)\) and \(P(t)\) denote the multiplicities of token \(t\in\mathcal{V}\). Their cardinalities are defined by
\begin{equation}
|G|=\sum_{t\in\mathcal{V}} G(t),
\qquad
|P|=\sum_{t\in\mathcal{V}} P(t).
\end{equation}

The \texttt{\textbf{set-match}} criterion is defined as
\begin{equation}
\mathrm{set\text{-}match}(G,P)=\mathbb{1}\bigl[G=P\ \wedge\ |G|>0\bigr]\in\{0,1\}.
\end{equation}
That is, the predicted multiset must match the gold multiset exactly, including token identities and multiplicities, and the gold multiset must be non-empty. Relative to \texttt{exact}, this criterion relaxes requirements on field order and surface-form realization, while still disallowing any omission, insertion, or multiplicity mismatch.

The \texttt{\textbf{overlap}} criterion is defined as
\begin{equation}
\mathrm{overlap}(G,P)=\frac{\sum_{t\in\mathcal{V}} \min\bigl(G(t),P(t)\bigr)}{\max\bigl(|G|,|P|\bigr)}\in[0,1].
\end{equation}
Here, the numerator is the total multiset intersection size, and the denominator is the larger of the two multiset cardinalities. Consequently, the metric penalizes both missing tokens and spurious tokens. We additionally define the boundary cases by
\begin{equation}
\mathrm{overlap}(\varnothing,\varnothing)=1,
\qquad
\mathrm{overlap}(G,P)=0 \ \text{if exactly one of } G,P \text{ is empty}.
\end{equation}

Under these definitions, the criteria satisfy the following implication relation:
\begin{equation}
\texttt{exact}\ \Longrightarrow\ \texttt{set-match}\ \Longrightarrow\ \mathrm{overlap}(G,P)=1.
\end{equation}
In other words, every prediction that is correct under \texttt{exact} is necessarily correct under \texttt{set-match}, and every prediction that satisfies \texttt{set-match} attains the maximal \texttt{overlap} score.

\section{Failure Mode Definition}

We define a dedicated error category for nearly every failure pattern that can be identified at the level of individual samples. These categories encompass \textit{Format Violation}, which also includes certain errors that may be interpreted as hallucinations, as well as canonical failure modes associated with color prediction, category recognition, text reading, counting, and order-sensitive spatial reasoning. Through this taxonomy of error types, we seek to enable a more fine-grained attribution of the failure mechanisms exhibited by VLMs. The formal definitions of these failure modes are provided below.

\paragraph{Refusal Failure(REF)} The model proactively declines to answer and fails to provide any parsable structured output. The generated response contains refusal-related phrases, such as ``I cannot'', ``I'm sorry'', ``as an AI'', ``unable to'', ``I can't help'', and ``unrecognizable'', and no valid JSON object can be extracted from the output (i.e., the parser returns \texttt{None}). Typical examples include cases in which the model refuses on the grounds of safety or capability, such as ``I'm sorry, but I can't identify text in images.'', or declines because the image is considered unclear, as in ``The image is too blurry to determine\ldots''

\paragraph{Format Violation (FMT)} The model attempts to provide an answer but fails to follow the required JSON schema, resulting in output that cannot be parsed. Specifically, no valid JSON object can be extracted from the response, but the case does not qualify as a refusal, as it contains no refusal-related phrases. This category reflects a failure of instruction following rather than a failure of perception or reasoning. Typical manifestations include responses written entirely in free-form natural language, or outputs wrapped in Markdown in which the enclosed JSON is malformed, incomplete, contains comments, or uses incorrect key names.

\paragraph{Misclassification (MSC)} In closed-set classification tasks, the JSON output is successfully parsed, but the predicted class is incorrect. This category applies to tasks such as LTR, SHP, ANM, and BLK, indicating that the output is structurally valid and schema-compliant, while the value of the \texttt{category} field (or its equivalent) does not match the ground truth. Representative errors include cases such as identifying ``B'' as ``D'' or recognizing ``triangle'' as ``pentagon''.

\paragraph{Color Confusion (CLR)} In tasks involving color binding or color selection, specifically CBL, the category may be identified correctly while the color field is predicted incorrectly. The JSON output is successfully parsed, but the value of the \texttt{color} field does not match the ground truth. For example, the model may correctly identify the letter as ``A'' while incorrectly reporting ``color'': ``orange'' for a red letter \texttt{A}.

\paragraph{OCR Error (OCE)} In scene text recognition tasks, the JSON output is successfully parsed, but the recognized word or string is incorrect. This category applies to tasks such as TXT. Specifically, the JSON output is valid, but the value of the \texttt{word} field does not exactly match the ground-truth string, either in a case-sensitive comparison or after the prescribed normalization procedure. Typical errors include omitted letters, inserted letters, character substitutions, and similar string-level mismatches.

\paragraph{Numeracy Error (NUM)} In tasks involving quantity estimation or object counting, the JSON output is successfully parsed, but the numerical field is incorrect. This category applies to tasks such as \texttt{CNT} (Count, total object counting), \texttt{BLR} (Blur Counting, counting under blurred conditions), and \texttt{TCT} (Targeted Count, counting after attribute-based filtering, e.g., ``the number of targets containing the letter X''). The JSON output is valid and schema-compliant, but the reported count does not match the ground truth.

\paragraph{Sequence Error (SEQ)} In tasks involving ordered lists or multi-target combinations, the JSON output is successfully parsed, but the content or ordering of the list does not match the ground truth. This category applies to tasks such as \texttt{CHR}, \texttt{CMP}, and \texttt{TRD}. Although the JSON output is valid and schema-compliant, the predicted \texttt{list} differs from the ground truth in either the set of elements, their order, or both. Representative errors include cases in which all elements are correct but appear in the wrong order, such as \texttt{["A","B","C"]} instead of the ground-truth \texttt{["B","A","C"]}.

\begin{table}[H]\centering\small
\setlength{\tabcolsep}{4pt}
\begin{adjustbox}{max width=\linewidth}
\begin{tabular}{lrrrrrrrr}
\toprule
 & REF & FMT & MSC & CLR & OCE & NUM & SEQ & TOT \\
\midrule
InternVL3\_5-1B-Flash & \cellcolor[HTML]{FFF5F0} 0.00 & \cellcolor[HTML]{FCA082} 6.70 & \cellcolor[HTML]{F44F39} \textcolor{white}{11.48} & \cellcolor[HTML]{FFF4EF} 0.12 & \cellcolor[HTML]{FEE8DE} 1.51 & \cellcolor[HTML]{BC141A} \textcolor{white}{15.94} & \cellcolor[HTML]{E43027} \textcolor{white}{13.24} & \cellcolor[HTML]{777777} \textcolor{white}{48.99} \\
InternVL3\_5-4B-Flash & \cellcolor[HTML]{FFF5F0} 0.00 & \cellcolor[HTML]{FDD5C4} 3.28 & \cellcolor[HTML]{F75C41} \textcolor{white}{10.78} & \cellcolor[HTML]{FFF1EA} 0.54 & \cellcolor[HTML]{FEE8DE} 1.55 & \cellcolor[HTML]{C7171C} \textcolor{white}{15.28} & \cellcolor[HTML]{CF1C1F} \textcolor{white}{14.75} & \cellcolor[HTML]{818181} \textcolor{white}{46.18} \\
InternVL3\_5-2B-Flash & \cellcolor[HTML]{FFF5F0} 0.00 & \cellcolor[HTML]{FEE0D2} 2.54 & \cellcolor[HTML]{FA6648} \textcolor{white}{10.22} & \cellcolor[HTML]{FFF4EE} 0.21 & \cellcolor[HTML]{FEE9DF} 1.45 & \cellcolor[HTML]{C8171C} \textcolor{white}{15.17} & \cellcolor[HTML]{C2161B} \textcolor{white}{15.60} & \cellcolor[HTML]{848484} \textcolor{white}{45.19} \\
Qwen3-VL-2B-Instruct & \cellcolor[HTML]{FFF5F0} 0.01 & \cellcolor[HTML]{FEEAE0} 1.39 & \cellcolor[HTML]{FB7252} 9.53 & \cellcolor[HTML]{FFF0E8} 0.69 & \cellcolor[HTML]{FFEBE2} 1.21 & \cellcolor[HTML]{BE151A} \textcolor{white}{15.79} & \cellcolor[HTML]{BB141A} \textcolor{white}{16.07} & \cellcolor[HTML]{858585} \textcolor{white}{44.69} \\
gemma-4-E2B-it & \cellcolor[HTML]{FFF5F0} 0.00 & \cellcolor[HTML]{FFF2EC} 0.34 & \cellcolor[HTML]{FC9576} 7.32 & \cellcolor[HTML]{FFF2EC} 0.32 & \cellcolor[HTML]{FFECE3} 1.11 & \cellcolor[HTML]{9C0D14} \textcolor{white}{17.88} & \cellcolor[HTML]{9F0E14} \textcolor{white}{17.68} & \cellcolor[HTML]{868686} \textcolor{white}{44.65} \\
InternVL3\_5-14B-Flash & \cellcolor[HTML]{FFF5F0} 0.00 & \cellcolor[HTML]{FFF0E9} 0.60 & \cellcolor[HTML]{FB694A} \textcolor{white}{10.06} & \cellcolor[HTML]{FFEFE8} 0.78 & \cellcolor[HTML]{FEE8DE} 1.56 & \cellcolor[HTML]{E02C26} \textcolor{white}{13.55} & \cellcolor[HTML]{A81016} \textcolor{white}{17.33} & \cellcolor[HTML]{888888} \textcolor{white}{43.88} \\
gemma-4-E4B-it & \cellcolor[HTML]{FFF5F0} 0.00 & \cellcolor[HTML]{FFF5F0} 0.00 & \cellcolor[HTML]{FCA689} 6.32 & \cellcolor[HTML]{FFF4EE} 0.22 & \cellcolor[HTML]{FFECE3} 1.10 & \cellcolor[HTML]{8A0812} \textcolor{white}{18.53} & \cellcolor[HTML]{AD1117} \textcolor{white}{16.92} & \cellcolor[HTML]{8C8C8C} \textcolor{white}{43.09} \\
InternVL3\_5-8B-Flash & \cellcolor[HTML]{FFF5F0} 0.00 & \cellcolor[HTML]{FEE7DB} 1.77 & \cellcolor[HTML]{F85D42} \textcolor{white}{10.69} & \cellcolor[HTML]{FFF4EE} 0.16 & \cellcolor[HTML]{FEE8DE} 1.50 & \cellcolor[HTML]{DC2924} \textcolor{white}{13.77} & \cellcolor[HTML]{C8171C} \textcolor{white}{15.18} & \cellcolor[HTML]{8C8C8C} \textcolor{white}{43.07} \\
Qwen3-VL-30B-A3B-Instruct & \cellcolor[HTML]{FCA98C} 6.16 & \cellcolor[HTML]{FCC1A8} 4.67 & \cellcolor[HTML]{FDD7C6} 3.20 & \cellcolor[HTML]{FFF5F0} 0.06 & \cellcolor[HTML]{FFEDE5} 1.01 & \cellcolor[HTML]{DA2723} \textcolor{white}{13.97} & \cellcolor[HTML]{E43027} \textcolor{white}{13.25} & \cellcolor[HTML]{8E8E8E} 42.32 \\
InternVL3\_5-30B-A3B-Flash & \cellcolor[HTML]{FFF5F0} 0.00 & \cellcolor[HTML]{FFF0E9} 0.55 & \cellcolor[HTML]{FB7050} \textcolor{white}{9.66} & \cellcolor[HTML]{FFEFE8} 0.78 & \cellcolor[HTML]{FEE8DE} 1.53 & \cellcolor[HTML]{EC382B} \textcolor{white}{12.68} & \cellcolor[HTML]{B21218} \textcolor{white}{16.64} & \cellcolor[HTML]{909090} 41.84 \\
Qwen3-VL-8B-Instruct & \cellcolor[HTML]{FFF5F0} 0.00 & \cellcolor[HTML]{FC8B6B} 7.92 & \cellcolor[HTML]{FC9C7D} 6.92 & \cellcolor[HTML]{FFF5F0} 0.02 & \cellcolor[HTML]{FFECE3} 1.12 & \cellcolor[HTML]{EC382B} \textcolor{white}{12.68} & \cellcolor[HTML]{F96245} \textcolor{white}{10.42} & \cellcolor[HTML]{999999} 39.08 \\
Qwen3-VL-4B-Instruct & \cellcolor[HTML]{FFF5F0} 0.01 & \cellcolor[HTML]{FC9E80} 6.79 & \cellcolor[HTML]{FCC2AA} 4.58 & \cellcolor[HTML]{FFF5F0} 0.07 & \cellcolor[HTML]{FFEBE2} 1.24 & \cellcolor[HTML]{DD2A25} \textcolor{white}{13.73} & \cellcolor[HTML]{F5523A} \textcolor{white}{11.27} & \cellcolor[HTML]{9F9F9F} 37.69 \\
InternVL3\_5-38B-Flash & \cellcolor[HTML]{FFF5F0} 0.00 & \cellcolor[HTML]{FFEDE5} 0.95 & \cellcolor[HTML]{FC8F6F} 7.70 & \cellcolor[HTML]{FFF5F0} 0.06 & \cellcolor[HTML]{FEE9DF} 1.48 & \cellcolor[HTML]{F0402F} \textcolor{white}{12.24} & \cellcolor[HTML]{CA181D} \textcolor{white}{15.01} & \cellcolor[HTML]{A0A0A0} 37.44 \\
GLM-4.6V-Flash & \cellcolor[HTML]{FFF5F0} 0.00 & \cellcolor[HTML]{FCBEA5} 4.79 & \cellcolor[HTML]{FCAE92} 5.80 & \cellcolor[HTML]{FFF2EB} 0.41 & \cellcolor[HTML]{FEEAE1} 1.27 & \cellcolor[HTML]{FC9D7F} 6.85 & \cellcolor[HTML]{D92523} \textcolor{white}{14.01} & \cellcolor[HTML]{B0B0B0} 33.13 \\
gpt-4o & \cellcolor[HTML]{FFF5F0} 0.05 & \cellcolor[HTML]{FFF5F0} 0.00 & \cellcolor[HTML]{FCBBA1} 5.01 & \cellcolor[HTML]{FFF5F0} 0.02 & \cellcolor[HTML]{FFEDE5} 0.94 & \cellcolor[HTML]{FB6C4C} \textcolor{white}{9.85} & \cellcolor[HTML]{DB2824} \textcolor{white}{13.85} & \cellcolor[HTML]{BEBEBE} 29.72 \\
gemma-4-26B-A4B-it & \cellcolor[HTML]{FEE4D8} 2.04 & \cellcolor[HTML]{FFF5F0} 0.04 & \cellcolor[HTML]{FEE1D4} 2.41 & \cellcolor[HTML]{FFF5F0} 0.04 & \cellcolor[HTML]{FFEEE6} 0.86 & \cellcolor[HTML]{FB7252} 9.50 & \cellcolor[HTML]{DD2A25} \textcolor{white}{13.68} & \cellcolor[HTML]{C1C1C1} 28.57 \\
gemma-4-31B-it & \cellcolor[HTML]{FDD5C4} 3.24 & \cellcolor[HTML]{FFF4EE} 0.18 & \cellcolor[HTML]{FEE6DA} 1.84 & \cellcolor[HTML]{FFF5F0} 0.00 & \cellcolor[HTML]{FFF1EA} 0.51 & \cellcolor[HTML]{FC8666} 8.26 & \cellcolor[HTML]{F44F39} \textcolor{white}{11.46} & \cellcolor[HTML]{CACACA} 25.49 \\
gemini-2.5-flash-preview & \cellcolor[HTML]{FFF5F0} 0.02 & \cellcolor[HTML]{FCB398} 5.54 & \cellcolor[HTML]{FDD7C6} 3.20 & \cellcolor[HTML]{FFF5F0} 0.04 & \cellcolor[HTML]{FFEEE7} 0.83 & \cellcolor[HTML]{FDC6B0} 4.24 & \cellcolor[HTML]{F4503A} \textcolor{white}{11.34} & \cellcolor[HTML]{CBCBCB} 25.21 \\
\bottomrule
\end{tabular}
\end{adjustbox}
\captionsetup{skip=6pt} 
\caption{Failure-mode rate (\%) per model (denominator = \#samples for that model). TOT $= 1 - \mathrm{Acc}_{\text{exact}}$ (\%); rows are sorted by TOT (descending). TOT=FAIL\_TOTAL}
\label{tab:fm_model}
\end{table}

As shown in Table~\ref{tab:fm_model}, \texttt{Gemini-2.5-Flash} and \texttt{Gemma-4-31B} achieve the lowest overall failure rates, both at approximately \(25\%\). They are followed closely by \texttt{GPT-4o} and \texttt{Gemma-4-26B}, whereas smaller models, such as \texttt{InternVL3.5-1B}, are concentrated above \(45\%\). Across all models, \textit{Numeracy Error} and \textit{Sequence Error} constitute the dominant sources of failure. Notably, even the strongest models are unable to reduce \textit{Sequence Error} below \(10\%\), suggesting that fine-grained compositional reasoning in tasks such as \texttt{CHR}, \texttt{CMP}, and \texttt{TRD} remains a shared bottleneck even for current state-of-the-art systems. By contrast, \textit{Color Confusion} and \textit{OCR Error} remain consistently negligible across models and therefore provide limited discriminative value.

A more fine-grained comparison further reveals that smaller models, including \texttt{InternVL3.5-1B/2B/4B} and \texttt{Qwen3-VL-2B}, exhibit substantially elevated \textit{Misclassification} rates, indicating that even closed-set category recognition remains unreliable at this scale. By contrast, mid-sized and larger models have largely mastered such classification tasks, and their residual failures are concentrated almost entirely in counting and order-sensitive enumeration. The \texttt{Qwen3-VL} family displays a distinct pattern of instruction-alignment weakness, characterized by unusually high \textit{Format Violation} rates; moreover, \texttt{Qwen3-VL-30B-A3B} uniquely exhibits a \textit{Refusal} rate of approximately \(6\%\), causing its overall failure rate to exceed those of its own \(4\)B and \(8\)B variants. Importantly, its relatively low \textit{Misclassification} rate suggests that this degradation is driven less by perceptual limitations than by an inability or unwillingness to comply with the required output format. In contrast, the \texttt{Gemma-4} and \texttt{InternVL3.5} families exhibit the cleanest scaling trends, indicating that they are the two model families for which performance gains from scaling are the most stable and systematic.

\section{Experimental Settings}

\subsection{Benchmark composition}

FineSightBench is primarily designed to evaluate the capabilities of vision-language models in \textbf{pixel-level perception} and fine-grained reasoning. It comprises two complementary data generation paradigms: one is a controllable synthetic canvas with a white background, and the other is a \textit{text-in-the-wild} setting in which English words are overlaid onto real natural-scene images. The latter is mainly intended to examine model performance under realistic image content and background interference. Additionally, this paper also will publish a dataset called \textbf{FineSightBench-Large} for potential training purposes, which contains ten times more samples and is constructed using the same data generation logic and methodology.

All images are standardized to a resolution of 448 $\times$ 448 pixels. The difficulty is mainly controlled by the pixel size of the target objects, with sizes ranging over 4, 8, 12, 16, 24, 32, and 48 pixels. The benchmark is divided into two splits, namely \textit{perception} and \textit{reasoning}. The \textit{perception} split contains 4,200 samples covering 6 task categories, while the \textit{reasoning} split contains 3,920 samples covering 6 task categories, resulting in a total of 8,120 samples.

In FineSightBench-Large, the scale is increased by a factor of ten and the dataset is primarily intended for training. Specifically, the \textit{perception} split contains 42,000 samples across 6 task categories, while the \textit{reasoning} split contains 39,200 samples across 6 task categories, yielding a total of 81,200 samples.

The perception split is designed for single-target recognition and contains 4,200 samples in total. Each task comprises 700 samples, formed by 7 pixel-size levels with 100 samples per level. It includes six task categories: letter recognition (LTR), animal recognition (ANM), shape recognition (SHP), block recognition (BLK), color block recognition (CBL), and text recognition (TXT). The first five categories are constructed from the synthetic canvas setting, while TXT is synthesized from source images in the \textit{text-in-the-wild} setting.

\begin{figure}[H]
  \centering
  \includegraphics[width=1.0\linewidth]{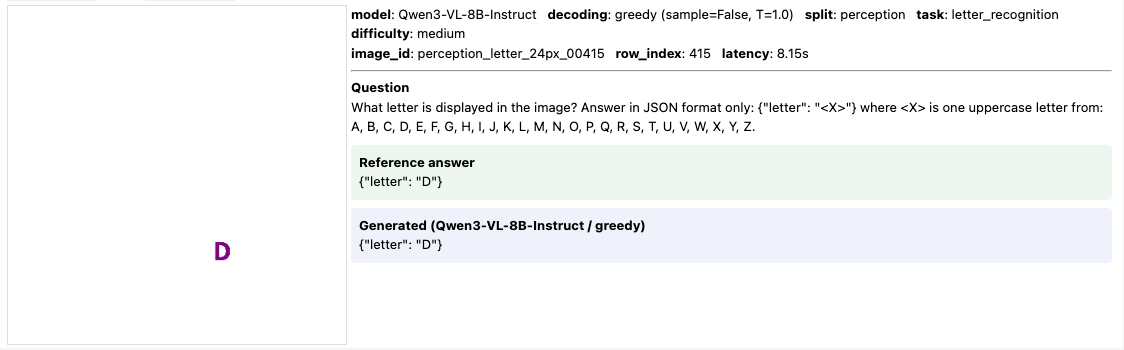}
  \caption{Example of the letter recognition (LTR) task in FineSightBench.}
  \label{fig:ltr_example}
\end{figure}

\begin{figure}[htbp]
  \centering
  \includegraphics[width=1.0\linewidth]{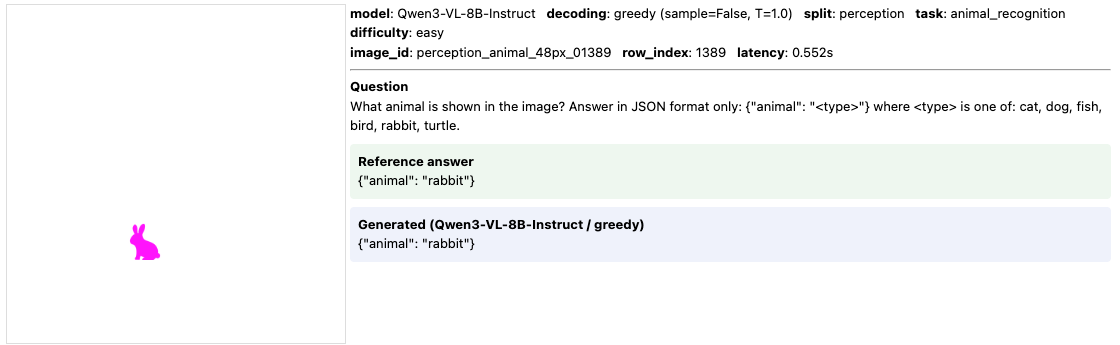}
  \caption{Example of the animal recognition (ANM) task in FineSightBench. We use animal icons rather than photographs of real animals because, when the target is reduced to a width of only 4--10 pixels, real images no longer preserve sufficient visual detail for reliable recognition. In contrast, icons derived from vector graphics allow precise control over target size while maintaining recognizable shape characteristics under extreme downscaling.}
  \label{fig:anm_example}
\end{figure}

\begin{figure}[htbp]
  \centering
  \includegraphics[width=1.0\linewidth]{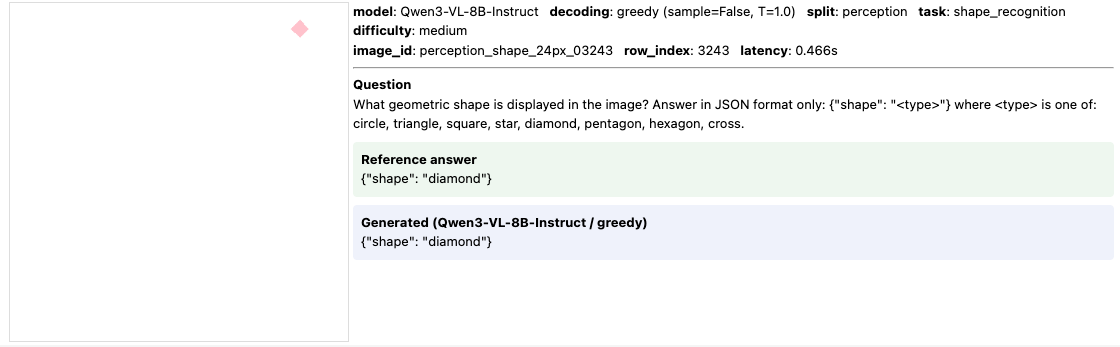}
  \caption{Example of the shape recognition (SHP) task in FineSightBench. This task evaluates whether the model can recognize the target shape, without requiring it to identify the target color. We also explicitly specify the candidate shape categories in the question prompt, as we observed that allowing vision-language models to produce unconstrained shape descriptions substantially increases the error rate for small size objects, whereas restricting the output space to predefined shape options leads to more reliable predictions.}
  \label{fig:shp_example}
\end{figure}

\begin{figure}[htbp]
  \centering
  \includegraphics[width=1.0\linewidth]{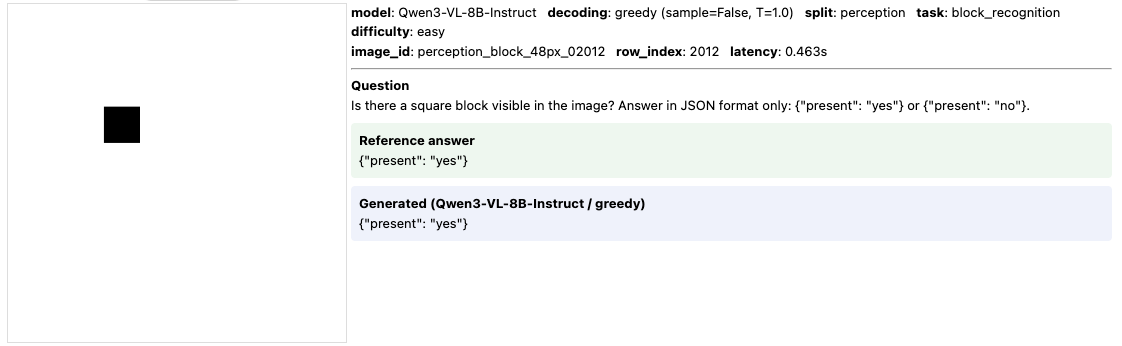}
  \caption{Example of the block recognition (BLK) task in FineSightBench. This is one of the simplest tasks in the benchmark, as it requires only direct visual detection of the target. The model is merely asked to determine whether the target is present, rather than to perform more complex reasoning such as shape inference or relational analysis.}
  \label{fig:blk_example}
\end{figure}

\begin{figure}[htbp]
  \centering
  \includegraphics[width=1.0\linewidth]{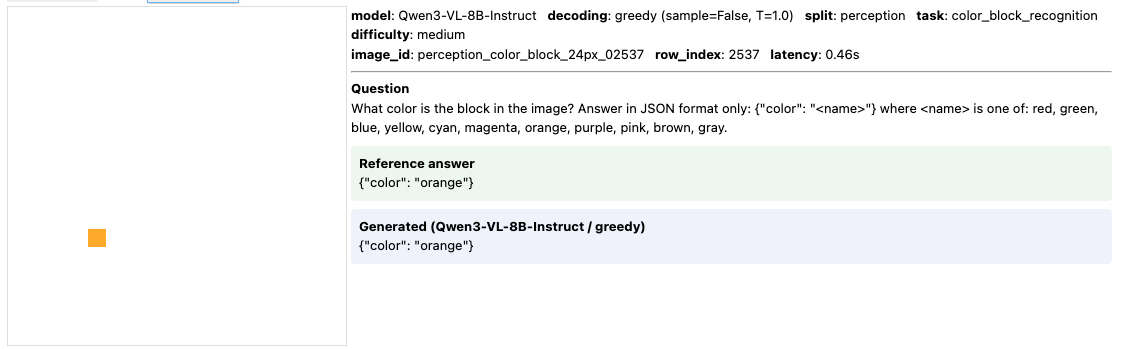}
  \caption{Example of the color block recognition (CBL) task in FineSightBench. This task is designed to examine whether a VLM can correctly perceive and identify color attributes, thereby probing its sensitivity to color information. Such color discrimination is also a fundamental component of human visual assessment.}
  \label{fig:cbl_example}
\end{figure}

\begin{figure}[htbp]
  \centering
  \includegraphics[width=1.0\linewidth]{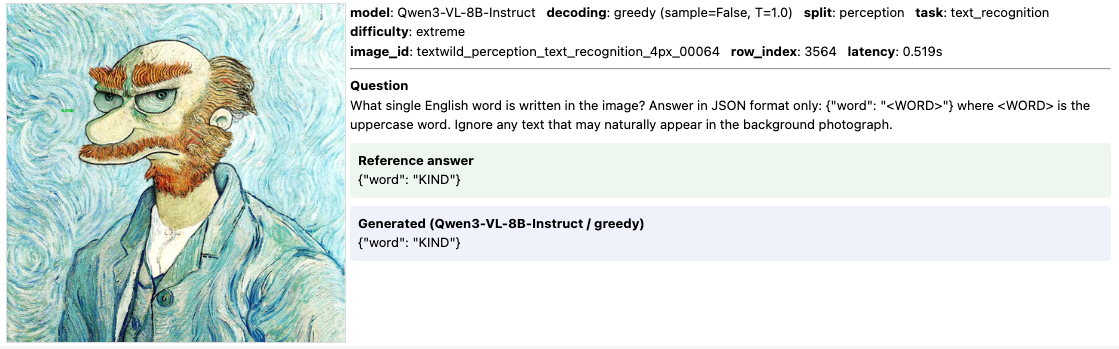}
  \caption{Example of the text recognition (TXT) task in FineSightBench. In this task, the model is required to detect a word embedded in a visually complex background and transcribe it correctly. Although the text may appear difficult to discern in the scaled figure, it becomes relatively clear when the image is viewed at its original resolution; nevertheless, recognition remains challenging because the surrounding background can introduce substantial visual interference.}
  \label{fig:txt_example}
\end{figure}

The reasoning split is designed to evaluate multi-target chain reasoning and contains 3,920 samples in total. It requires models to solve more complex compositional reasoning tasks, such as counting, ordering, and spatial relation understanding. This split includes six task categories: Chain Reasoning (CHR), Comparison Chain Reasoning (CMP), Counting Chain Reasoning (CNT), Blur Chain Reasoning (BLR), Text Reading (TRD), and Text Counting (TCT). The first four categories are constructed from the synthetic canvas setting, while the last two are synthesized from the \textit{text-in-the-wild} setting. The data generation code has been fully anonymized and made publicly available as open source.

\begin{figure}[htbp]
  \centering
  \includegraphics[width=1.0\linewidth]{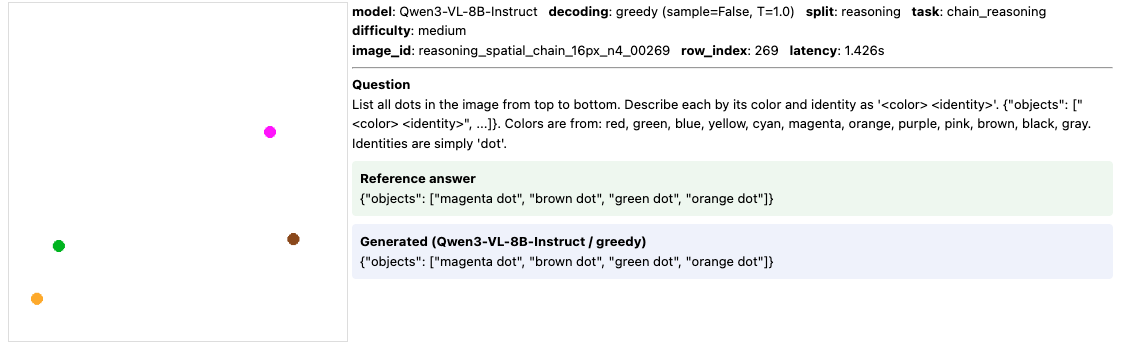}
  \caption{Example of the chain reasoning (CHR) task in FineSightBench. The model is required to enumerate all targets in the image in a specified order, such as from top to bottom or from left to right, and report both their colors and categories. The prompt explicitly specifies the candidate color set and target categories to reduce output ambiguity and improve evaluation consistency.}
  \label{fig:chr_example}
\end{figure}

\begin{figure}[htbp]
  \centering
  \includegraphics[width=1.0\linewidth]{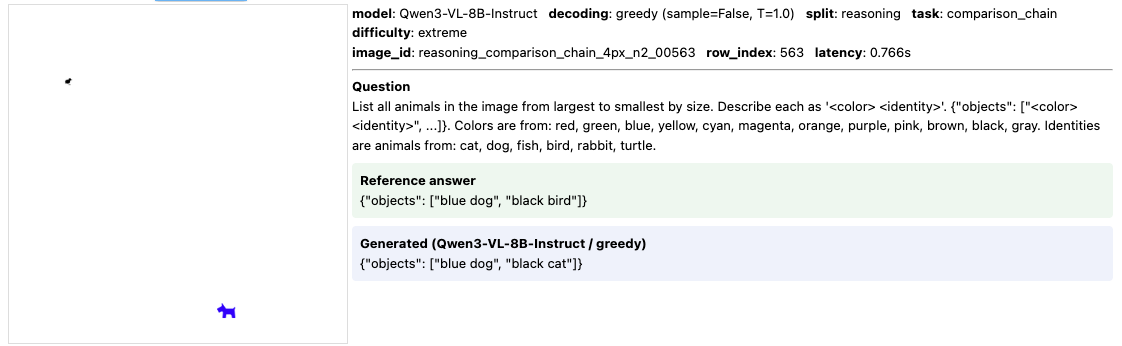}
  \caption{Example of the comparison chain reasoning (CMP) task in FineSightBench. Fine-grained size comparison is a critical component of this task, making it particularly challenging. The model is required not only to rank the targets in ascending or descending order of size, but also to identify the corresponding object or animal category and its color for each target.}
  \label{fig:cmp_example}
\end{figure}

\begin{figure}[htbp]
  \centering
  \includegraphics[width=1.0\linewidth]{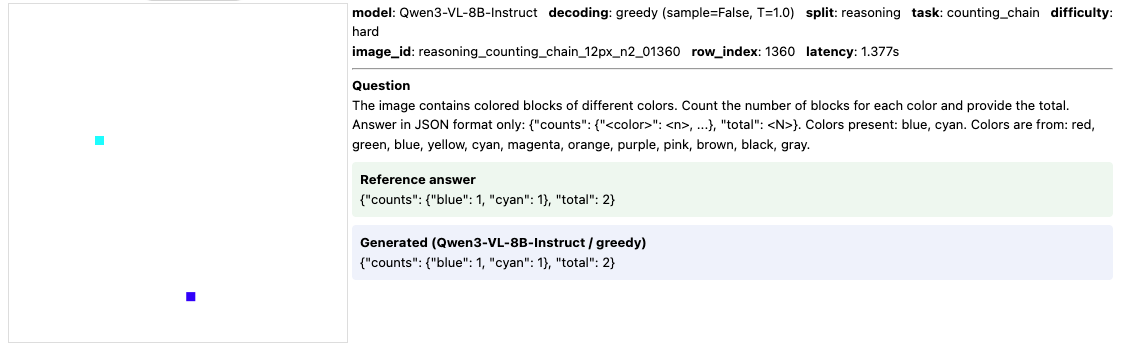}
  \caption{Example of the counting chain reasoning (CNT) task in FineSightBench. This task requires the model to perform category-wise counting for different types of targets and then derive the total number of objects in the image through multi-step reasoning.}
  \label{fig:cnt_example}
\end{figure}

\begin{figure}[htbp]
  \centering
  \includegraphics[width=1.0\linewidth]{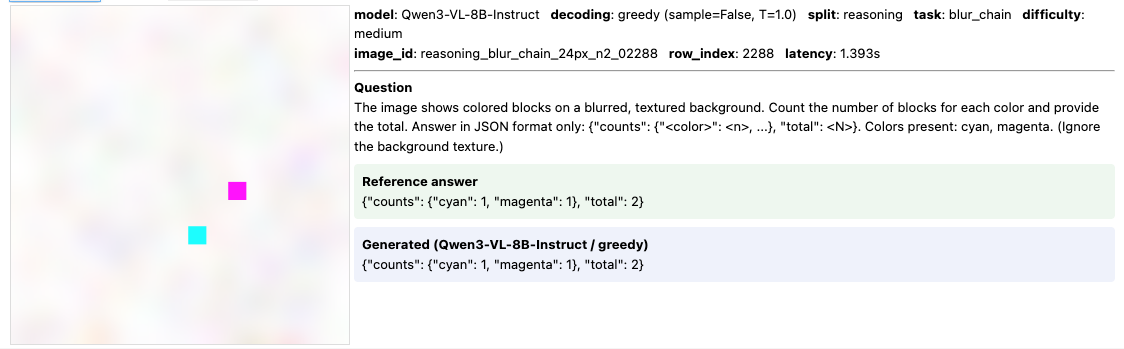}
  \caption{Example of the blur chain reasoning (BLR) task in FineSightBench. In this task, we introduce artificial visual interference by adding colored patches to the background and applying blur, in order to evaluate the accuracy and robustness of vision-language models under controlled perturbations.}
  \label{fig:blr_example}
\end{figure}

\begin{figure}[htbp]
  \centering
  \includegraphics[width=1.0\linewidth]{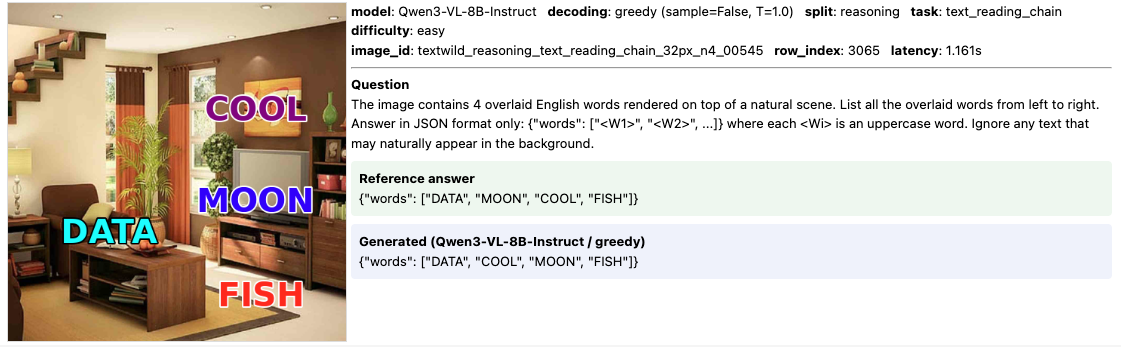}
  \caption{Example of the text reading (TRD) task in FineSightBench. The model is required to detect the words in the image and output them in a predefined spatial order, such as from left to right or from top to bottom.}
  \label{fig:trd_example}
\end{figure}

\begin{figure}[htbp]
  \centering
  \includegraphics[width=1.0\linewidth]{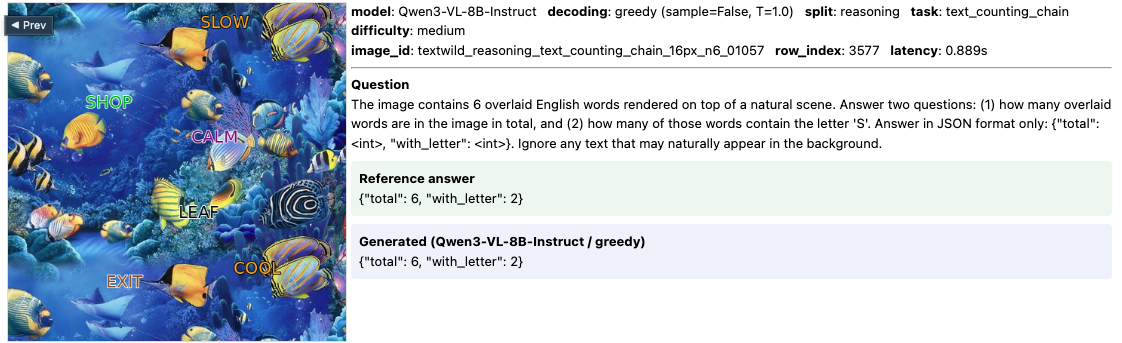}
  \caption{Example of the text counting (TCT) task in FineSightBench. This task focuses on counting as well as rule-conditioned counting. For instance, the model may be asked to count the number of words that satisfy a specific condition, such as containing the letter ``S''.}
  \label{fig:tct_example}
\end{figure}

\subsection{Inference Settings}

For each model, we test every sample using three decoding settings to examine how decoding strategy affects performance: deterministic decoding without sampling, sampling with a low temperature of 0.1, and sampling with a temperature of 1.0. The maximum generation length is fixed to \texttt{max\_new\_tokens=1024} for all models in ModelSpec. Other decoding hyperparameters follow the default settings, including \texttt{top\_p=1.0}, \texttt{top\_k=50}, \texttt{num\_beams=1}, and \texttt{repetition\_penalty=1.0}. In practice, we do not place particular emphasis on these parameters, as recent findings suggest that temperature is the dominant factor affecting decoding behavior \cite{LI2025242}. For model inputs, all images are uniformly converted to RGB using \texttt{convert("RGB")}, and the prompt is taken directly from the dataset question field, without any system prompt or few-shot exemplars. The evaluated models are listed in Table \ref{tab:vlm_selected_models} and cover most major open-source vision-language models. Among them, GLM-4.6V-Flash is noteworthy for its support for textual reasoning. To minimize implementation-related variation, all models are evaluated in \texttt{bfloat16} precision within a unified inference framework based on the Hugging Face Transformers library.

\begin{figure}[htbp]
  \centering
  \includegraphics[width=1.0\linewidth]{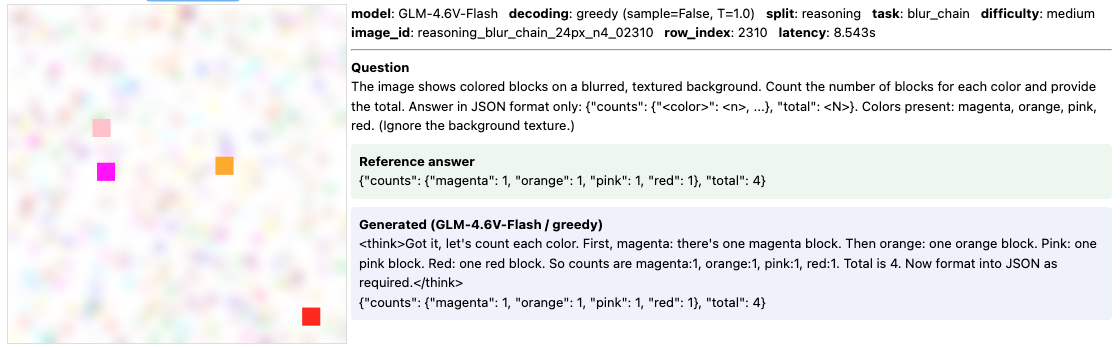}
  \caption{Example outputs from GLM-4.6V-Flash. As a reasoning-oriented model, it typically generates an explicit thinking process before providing the final answer, which leads to an output format different from that of standard chat models. This characteristic offers a more interpretable view of how the vision-language model arrives at its prediction.}
  \label{fig:glm_example}
\end{figure}


\definecolor{groupbg}{gray}{0.90}
\definecolor{subbg}{gray}{0.96}

\begin{table*}[htbp]
\centering
\small
\setlength{\tabcolsep}{6pt}
\renewcommand{\arraystretch}{1.25}
\begin{adjustbox}{max width=\textwidth}
\begin{tabular}{p{6.2cm} p{2.4cm} p{1.4cm} p{3.2cm}}
\toprule
\textbf{Model} & \textbf{Total Size} & \textbf{Release} & \textbf{Reference} \\
\midrule

\rowcolor{groupbg}
\multicolumn{4}{l}{\textbf{Open-source / Open-weight Models}} \\

\rowcolor{subbg}
\multicolumn{4}{l}{\quad\textit{Alibaba Qwen}} \\
\quad Qwen3-VL-2B-Instruct         & 2B          & 2025-10 & \cite{qwen3_vl_2b_instruct_hf} \\
\quad Qwen3-VL-4B-Instruct         & 4B          & 2025-10 & \cite{qwen3_vl_4b_instruct_hf} \\
\quad Qwen3-VL-8B-Instruct         & 8B          & 2025-10 & \cite{qwen3_vl_8b_instruct_hf} \\
\quad Qwen3-VL-30B-A3B-Instruct    & 30B-A3B     & 2025-10 & \cite{qwen3_vl_30b_a3b_instruct_hf} \\


\rowcolor{subbg}
\multicolumn{4}{l}{\quad\textit{OpenGVLab}} \\
\quad InternVL3\_5-1B-Flash        & 1B          & 2025-08 & \cite{internvl35_1b_flash_hf} \\
\quad InternVL3\_5-2B-Flash        & 2B          & 2025-08 & \cite{internvl35_2b_flash_hf} \\
\quad InternVL3\_5-4B-Flash        & 4B          & 2025-08 & \cite{internvl35_4b_flash_hf} \\
\quad InternVL3\_5-8B-Flash        & 8B          & 2025-08 & \cite{internvl35_8b_flash_hf} \\
\quad InternVL3\_5-14B-Flash       & 15.1B       & 2025-08 & \cite{internvl35_14b_flash_hf} \\
\quad InternVL3\_5-30B-A3B-Flash   & 30.8B-A3B   & 2025-08 & \cite{internvl35_30b_a3b_flash_hf} \\
\quad InternVL3\_5-38B-Flash       & 38.4B       & 2025-08 & \cite{internvl35_38b_flash_hf} \\


\rowcolor{subbg}
\multicolumn{4}{l}{\quad\textit{Z.ai}} \\
\quad GLM-4.6V-Flash               & 9B          & 2025-12 & \cite{glm46v_flash_hf} \\

\rowcolor{subbg}
\multicolumn{4}{l}{\quad\textit{Google}} \\
\quad gemma-4-E2B-it               & 5B          & 2026-04 & \cite{gemma4_e2b_it_hf} \\
\quad gemma-4-E4B-it                   & 8B        & 2026-04 & \cite{gemma4_e4b_it_hf} \\
\quad gemma-4-26B-A4B-it               & 27B          & 2026-04 & \cite{gemma4_26b_a4b_it_hf} \\
\quad gemma-4-31B-it                  & 33B        & 2026-04 & \cite{gemma4_31b_it_hf} \\

\midrule

\rowcolor{groupbg}
\multicolumn{4}{l}{\textbf{Closed-source Models}} \\

\quad GPT-4o                       & Undisclosed & 2024-05 & \cite{gpt4o_openai} \\

\quad gemini-2.5-flash-preview              & Undisclosed & 2025-12 & \cite{gemini3_flash_preview} \\

\bottomrule
\end{tabular}
\end{adjustbox}
\caption{
  Selected vision-language models (VLMs), organized by provider and grouped into
  open-source/open-weight and closed-source categories.
  ``Active'' denotes activated parameters in mixture-of-experts (MoE) architectures.
  ``Undisclosed'' indicates proprietary or officially unpublished specifications.
}
\label{tab:vlm_selected_models}
\end{table*}

\section{Computational Resource}

Because each question is evaluated three times under different decoding settings, the final evaluation produces a total of 438,480 outputs. For GPT-4o and Gemini-2.5-Flash-Preview, we are unable to accurately determine the corresponding GPU and CPU resource consumption. However, for most of the other open-source models, inference was conducted on four NVIDIA H100 GPUs, providing a total of \(4 \times 80\,\mathrm{GB} = 320\,\mathrm{GB}\) of VRAM. Overall, the experiments required approximately 3,200 GPU hours for a single decoding strategy, and about 10,000 GPU hours in total across all decoding settings.




\section{Oracle Perception Ablation}

An additional research question of particular interest is that \textit{perception} and \textit{reasoning} are not truly disentangled in the current setting. Especially for reasoning-oriented examples, it is often unclear whether a VLM fails because it cannot correctly perceive the relevant visual information, or because it has perceived the information but fails to carry out the subsequent textual reasoning. To further investigate this issue, we introduce an \textit{oracle setting}, in which the model is provided with the gold object list or gold recognized tokens, so that it bypasses the visual perception stage and only performs the downstream reasoning operations, such as sorting, comparison, and counting. Specifically, for each reasoning example, we directly provide the manually annotated and fully correct visual perception outputs, namely the fine-grained metadata including object identities, colors, positions, sizes, and textual content. Under this setting, the VLM is only required to perform the final reasoning step, such as counting, sorting, and comparison.

We evaluate \textit{Qwen3-VL-2B-Instruct} and \textit{Qwen3-VL-4B-Instruct} on our dataset, using all examples in the \textit{Reasoning} split and adopting greedy decoding throughout. Overall, we observe a performance gap of \(10.4\%\) on \textit{Qwen3-VL-2B-Instruct} between the oracle setting and the original setting, where the former removes the need for target localization and recognition and only requires reasoning, while the latter requires the model to perform both perception and reasoning. On \textit{Qwen3-VL-4B-Instruct}, this gap increases to \(22.6\%\).

These results suggest that, once the overall model capacity is sufficiently strong, the primary challenge for VLMs on complex reasoning tasks involving small-pixel targets lies more in \textit{seeing} than in \textit{reasoning}. In other words, the main bottleneck is often whether the model can accurately perceive the relevant visual content, rather than whether it can execute the reasoning process itself. Only when the model can see the target clearly can it reliably solve the downstream reasoning task. In our view, much of the reasoning capability in VLMs is inherited from their underlying LLMs, whose reasoning ability has already been validated on a wide range of benchmarks. Therefore, for such tasks, accurately translating visual inputs into faithful textual or structured representations, and then reasoning over them, may be more critical than reasoning alone.

\begin{figure}[htbp]
  \centering
  \includegraphics[width=1.0\linewidth]{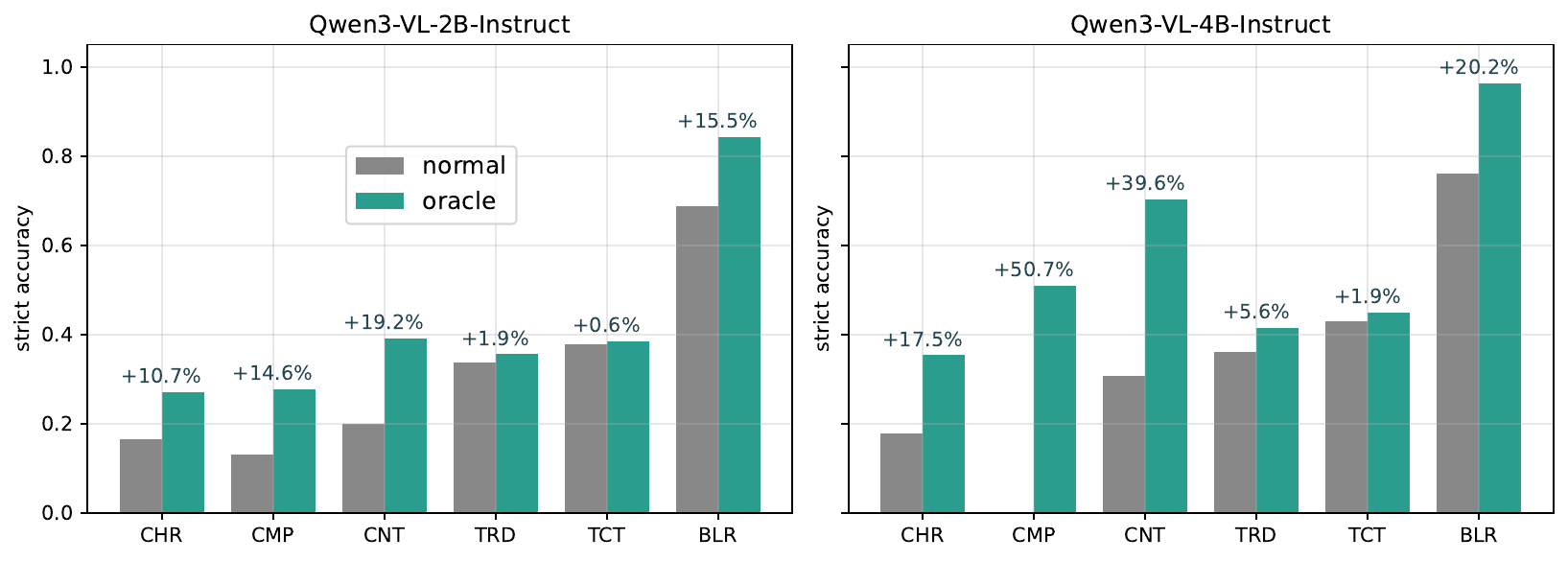}
  \caption{Normal vs Oracle perception — strict accuracy per reasoning task}
  \label{fig:norm_vs_oracle}
\end{figure}

As shown in Figure~\ref{fig:norm_vs_oracle}, we further present the performance gains across different tasks. We observe that when all specific information about the target in the image is provided to the model, performance improves across all tasks for both models. Naturally, the magnitude of improvement varies by task; with the exception of TRD and TCT, the performance gains for the remaining tasks all exceed $10\%$. Although we evaluate only two models, the overall trend suggests that VLMs are strong in reasoning, particularly text-based reasoning. However, their capability in the full pipeline from visual perception to reasoning remains limited. This indicates that detecting and understanding small pixel-level targets remains a notable weakness of VLMs.

\section{Additional Results}

\definecolor{cellbg0}{RGB}{0,104,55}
\definecolor{cellbg1}{RGB}{228,76,52}
\definecolor{cellbg2}{RGB}{247,129,76}
\definecolor{cellbg3}{RGB}{165,0,38}
\definecolor{cellbg4}{RGB}{179,13,38}
\definecolor{cellbg5}{RGB}{253,187,108}
\definecolor{cellbg6}{RGB}{227,73,51}
\definecolor{cellbg7}{RGB}{214,47,39}
\definecolor{cellbg8}{RGB}{51,164,86}
\definecolor{cellbg9}{RGB}{251,253,186}
\definecolor{cellbg10}{RGB}{253,254,188}
\definecolor{cellbg11}{RGB}{254,218,134}
\definecolor{cellbg12}{RGB}{203,233,130}
\definecolor{cellbg13}{RGB}{254,214,131}
\definecolor{cellbg14}{RGB}{251,163,92}
\definecolor{cellbg15}{RGB}{11,125,66}
\definecolor{cellbg16}{RGB}{177,222,113}
\definecolor{cellbg17}{RGB}{25,151,80}
\definecolor{cellbg18}{RGB}{14,130,69}
\definecolor{cellbg19}{RGB}{63,170,89}
\definecolor{cellbg20}{RGB}{18,138,73}
\definecolor{cellbg21}{RGB}{81,179,94}
\definecolor{cellbg22}{RGB}{192,26,39}
\definecolor{cellbg23}{RGB}{247,132,78}
\definecolor{cellbg24}{RGB}{225,68,48}
\definecolor{cellbg25}{RGB}{218,240,141}
\definecolor{cellbg26}{RGB}{221,61,45}
\definecolor{cellbg27}{RGB}{190,24,39}
\definecolor{cellbg28}{RGB}{33,156,82}
\definecolor{cellbg29}{RGB}{185,225,118}
\definecolor{cellbg30}{RGB}{3,110,58}
\definecolor{cellbg31}{RGB}{223,242,147}
\definecolor{cellbg32}{RGB}{120,197,101}
\definecolor{cellbg33}{RGB}{125,199,101}
\definecolor{cellbg34}{RGB}{255,247,178}
\definecolor{cellbg35}{RGB}{183,224,117}
\definecolor{cellbg36}{RGB}{110,192,100}
\definecolor{cellbg37}{RGB}{45,161,85}
\definecolor{cellbg38}{RGB}{142,207,103}
\definecolor{cellbg39}{RGB}{171,219,109}
\definecolor{cellbg40}{RGB}{60,169,89}
\definecolor{cellbg41}{RGB}{42,160,84}
\definecolor{cellbg42}{RGB}{187,226,120}
\definecolor{cellbg43}{RGB}{152,211,104}
\definecolor{cellbg44}{RGB}{21,144,76}
\definecolor{cellbg45}{RGB}{36,157,83}
\definecolor{cellbg46}{RGB}{30,154,81}
\definecolor{cellbg47}{RGB}{254,222,137}
\definecolor{cellbg48}{RGB}{242,104,65}
\definecolor{cellbg49}{RGB}{181,15,38}
\definecolor{cellbg50}{RGB}{224,66,47}
\definecolor{cellbg51}{RGB}{204,38,39}
\definecolor{cellbg52}{RGB}{254,231,151}
\definecolor{cellbg53}{RGB}{221,241,145}
\definecolor{cellbg54}{RGB}{7,117,62}
\definecolor{cellbg55}{RGB}{169,218,108}
\definecolor{cellbg56}{RGB}{75,176,92}
\definecolor{cellbg57}{RGB}{54,166,87}
\definecolor{cellbg58}{RGB}{245,119,72}
\definecolor{cellbg59}{RGB}{230,245,157}
\definecolor{cellbg60}{RGB}{248,134,79}
\definecolor{cellbg61}{RGB}{10,123,65}
\definecolor{cellbg62}{RGB}{1,106,56}
\definecolor{cellbg63}{RGB}{24,149,79}
\definecolor{cellbg64}{RGB}{16,134,71}
\definecolor{cellbg65}{RGB}{90,183,96}
\definecolor{cellbg66}{RGB}{213,237,136}
\definecolor{cellbg67}{RGB}{99,188,98}
\definecolor{cellbg68}{RGB}{255,246,176}
\definecolor{cellbg69}{RGB}{254,228,145}
\definecolor{cellbg70}{RGB}{224,242,149}
\definecolor{cellbg71}{RGB}{15,132,70}
\definecolor{cellbg72}{RGB}{17,136,72}
\definecolor{cellbg73}{RGB}{48,163,86}
\definecolor{cellbg74}{RGB}{238,97,62}
\definecolor{cellbg75}{RGB}{127,200,102}
\definecolor{cellbg76}{RGB}{4,112,59}

\begin{table}[htbp]
  \centering
  \caption{RT50 Of the main experiments. This reflect all the models in every task which pixel is the minimum acceptable pixel. Samller means the models has good visual capability on this tasks with higher than 50\% of accuracy.}
  \label{tab:rt50}
  \begin{adjustbox}{max width=\textwidth}
    \begin{tabular}{lrrrrrrrrrrrr}
    \toprule
    model & LTR & SHP & ANM & BLK & CBL & TXT & CHR & CMP & CNT & BLR & TRD & TCT \\
    \midrule
    GLM-4.6V-Flash & \cellcolor{cellbg0}\color{white}4.0 & \cellcolor{cellbg8}\color{black}5.7 & \cellcolor{cellbg21}\color{black}9.5 & \cellcolor{cellbg37}\color{black}6.1 & \cellcolor{cellbg0}\color{white}4.0 & \cellcolor{cellbg47}\color{black}6.1 & \cellcolor{cellbg3}\color{white}49.0 & \cellcolor{cellbg3}\color{white}49.0 & \cellcolor{cellbg57}\color{black}17.0 & \cellcolor{cellbg0}\color{white}4.0 & \cellcolor{cellbg65}\color{black}15.5 & \cellcolor{cellbg18}\color{white}7.2 \\
    InternVL3\_5-14B-Flash & \cellcolor{cellbg1}\color{black}5.6 & \cellcolor{cellbg9}\color{black}8.4 & \cellcolor{cellbg22}\color{white}20.8 & \cellcolor{cellbg32}\color{black}7.8 & \cellcolor{cellbg0}\color{white}4.0 & \cellcolor{cellbg3}\color{white}6.9 & \cellcolor{cellbg3}\color{white}49.0 & \cellcolor{cellbg3}\color{white}49.0 & \cellcolor{cellbg58}\color{black}40.9 & \cellcolor{cellbg44}\color{white}4.7 & \cellcolor{cellbg66}\color{black}24.0 & \cellcolor{cellbg17}\color{white}9.1 \\
    InternVL3\_5-1B-Flash & \cellcolor{cellbg2}\color{black}5.5 & \cellcolor{cellbg3}\color{white}12.4 & \cellcolor{cellbg3}\color{white}21.6 & \cellcolor{cellbg38}\color{black}8.4 & \cellcolor{cellbg0}\color{white}4.0 & \cellcolor{cellbg26}\color{white}6.6 & \cellcolor{cellbg3}\color{white}49.0 & \cellcolor{cellbg3}\color{white}49.0 & \cellcolor{cellbg3}\color{white}49.0 & \cellcolor{cellbg0}\color{white}4.0 & \cellcolor{cellbg3}\color{white}49.0 & \cellcolor{cellbg33}\color{black}15.2 \\
    InternVL3\_5-2B-Flash & \cellcolor{cellbg3}\color{white}5.9 & \cellcolor{cellbg10}\color{black}8.5 & \cellcolor{cellbg23}\color{black}18.2 & \cellcolor{cellbg39}\color{black}9.2 & \cellcolor{cellbg0}\color{white}4.0 & \cellcolor{cellbg48}\color{black}6.5 & \cellcolor{cellbg3}\color{white}49.0 & \cellcolor{cellbg3}\color{white}49.0 & \cellcolor{cellbg3}\color{white}49.0 & \cellcolor{cellbg0}\color{white}4.0 & \cellcolor{cellbg3}\color{white}49.0 & \cellcolor{cellbg19}\color{black}11.3 \\
    InternVL3\_5-30B-A3B-Flash & \cellcolor{cellbg4}\color{white}5.9 & \cellcolor{cellbg11}\color{black}9.4 & \cellcolor{cellbg24}\color{white}19.6 & \cellcolor{cellbg40}\color{black}6.4 & \cellcolor{cellbg0}\color{white}4.0 & \cellcolor{cellbg49}\color{white}6.8 & \cellcolor{cellbg3}\color{white}49.0 & \cellcolor{cellbg3}\color{white}49.0 & \cellcolor{cellbg59}\color{black}28.0 & \cellcolor{cellbg64}\color{white}4.5 & \cellcolor{cellbg42}\color{black}22.0 & \cellcolor{cellbg72}\color{white}7.7 \\
    InternVL3\_5-38B-Flash & \cellcolor{cellbg5}\color{black}5.3 & \cellcolor{cellbg12}\color{black}7.5 & \cellcolor{cellbg25}\color{black}12.9 & \cellcolor{cellbg41}\color{black}6.0 & \cellcolor{cellbg0}\color{white}4.0 & \cellcolor{cellbg50}\color{white}6.6 & \cellcolor{cellbg3}\color{white}49.0 & \cellcolor{cellbg3}\color{white}49.0 & \cellcolor{cellbg59}\color{black}28.0 & \cellcolor{cellbg28}\color{white}4.9 & \cellcolor{cellbg67}\color{black}16.0 & \cellcolor{cellbg71}\color{white}7.3 \\
    InternVL3\_5-4B-Flash & \cellcolor{cellbg6}\color{black}5.6 & \cellcolor{cellbg13}\color{black}9.5 & \cellcolor{cellbg26}\color{white}19.8 & \cellcolor{cellbg42}\color{black}9.7 & \cellcolor{cellbg0}\color{white}4.0 & \cellcolor{cellbg22}\color{white}6.8 & \cellcolor{cellbg3}\color{white}49.0 & \cellcolor{cellbg3}\color{white}49.0 & \cellcolor{cellbg3}\color{white}49.0 & \cellcolor{cellbg0}\color{white}4.0 & \cellcolor{cellbg68}\color{black}29.7 & \cellcolor{cellbg3}\color{white}49.0 \\
    InternVL3\_5-8B-Flash & \cellcolor{cellbg7}\color{white}5.7 & \cellcolor{cellbg14}\color{black}10.2 & \cellcolor{cellbg27}\color{white}20.8 & \cellcolor{cellbg43}\color{black}8.6 & \cellcolor{cellbg0}\color{white}4.0 & \cellcolor{cellbg51}\color{white}6.7 & \cellcolor{cellbg3}\color{white}49.0 & \cellcolor{cellbg3}\color{white}49.0 & \cellcolor{cellbg60}\color{black}40.0 & \cellcolor{cellbg0}\color{white}4.0 & \cellcolor{cellbg38}\color{black}18.7 & \cellcolor{cellbg28}\color{white}9.6 \\
    Qwen3-VL-2B-Instruct & \cellcolor{cellbg0}\color{white}4.0 & \cellcolor{cellbg15}\color{white}5.0 & \cellcolor{cellbg28}\color{white}8.6 & \cellcolor{cellbg3}\color{white}20.6 & \cellcolor{cellbg0}\color{white}4.0 & \cellcolor{cellbg34}\color{black}5.9 & \cellcolor{cellbg3}\color{white}49.0 & \cellcolor{cellbg3}\color{white}49.0 & \cellcolor{cellbg3}\color{white}49.0 & \cellcolor{cellbg0}\color{white}4.0 & \cellcolor{cellbg3}\color{white}49.0 & \cellcolor{cellbg3}\color{white}49.0 \\
    Qwen3-VL-30B-A3B-Instruct & \cellcolor{cellbg0}\color{white}4.0 & \cellcolor{cellbg16}\color{black}7.2 & \cellcolor{cellbg29}\color{black}11.9 & \cellcolor{cellbg44}\color{white}5.4 & \cellcolor{cellbg0}\color{white}4.0 & \cellcolor{cellbg25}\color{black}5.7 & \cellcolor{cellbg3}\color{white}49.0 & \cellcolor{cellbg3}\color{white}49.0 & \cellcolor{cellbg3}\color{white}49.0 & \cellcolor{cellbg0}\color{white}4.0 & \cellcolor{cellbg56}\color{black}14.7 & \cellcolor{cellbg73}\color{black}10.4 \\
    Qwen3-VL-4B-Instruct & \cellcolor{cellbg0}\color{white}4.0 & \cellcolor{cellbg15}\color{white}5.1 & \cellcolor{cellbg30}\color{white}7.1 & \cellcolor{cellbg45}\color{white}5.9 & \cellcolor{cellbg0}\color{white}4.0 & \cellcolor{cellbg52}\color{black}6.0 & \cellcolor{cellbg3}\color{white}49.0 & \cellcolor{cellbg3}\color{white}49.0 & \cellcolor{cellbg3}\color{white}49.0 & \cellcolor{cellbg0}\color{white}4.0 & \cellcolor{cellbg69}\color{black}32.0 & \cellcolor{cellbg74}\color{black}40.9 \\
    Qwen3-VL-8B-Instruct & \cellcolor{cellbg0}\color{white}4.0 & \cellcolor{cellbg17}\color{white}5.5 & \cellcolor{cellbg31}\color{black}13.0 & \cellcolor{cellbg28}\color{white}5.8 & \cellcolor{cellbg0}\color{white}4.0 & \cellcolor{cellbg53}\color{black}5.7 & \cellcolor{cellbg3}\color{white}49.0 & \cellcolor{cellbg3}\color{white}49.0 & \cellcolor{cellbg3}\color{white}49.0 & \cellcolor{cellbg0}\color{white}4.0 & \cellcolor{cellbg70}\color{black}25.1 & \cellcolor{cellbg75}\color{black}15.3 \\
    gemini-2.5-flash-preview & \cellcolor{cellbg0}\color{white}4.0 & \cellcolor{cellbg18}\color{white}5.1 & \cellcolor{cellbg0}\color{white}6.9 & \cellcolor{cellbg0}\color{white}4.0 & \cellcolor{cellbg0}\color{white}4.0 & \cellcolor{cellbg54}\color{white}4.9 & \cellcolor{cellbg3}\color{white}49.0 & \cellcolor{cellbg3}\color{white}49.0 & \cellcolor{cellbg0}\color{white}11.8 & \cellcolor{cellbg0}\color{white}4.0 & \cellcolor{cellbg71}\color{white}10.3 & \cellcolor{cellbg0}\color{white}4.7 \\
    gemma-4-26B-A4B-it & \cellcolor{cellbg0}\color{white}4.0 & \cellcolor{cellbg15}\color{white}5.0 & \cellcolor{cellbg32}\color{black}10.3 & \cellcolor{cellbg0}\color{white}4.0 & \cellcolor{cellbg0}\color{white}4.0 & \cellcolor{cellbg54}\color{white}4.9 & \cellcolor{cellbg45}\color{white}15.7 & \cellcolor{cellbg3}\color{white}49.0 & \cellcolor{cellbg61}\color{white}13.3 & \cellcolor{cellbg0}\color{white}4.0 & \cellcolor{cellbg17}\color{white}12.0 & \cellcolor{cellbg64}\color{white}7.4 \\
    gemma-4-31B-it & \cellcolor{cellbg0}\color{white}4.0 & \cellcolor{cellbg0}\color{white}4.7 & \cellcolor{cellbg33}\color{black}10.4 & \cellcolor{cellbg0}\color{white}4.0 & \cellcolor{cellbg0}\color{white}4.0 & \cellcolor{cellbg0}\color{white}4.9 & \cellcolor{cellbg0}\color{white}11.3 & \cellcolor{cellbg0}\color{white}14.5 & \cellcolor{cellbg62}\color{white}12.0 & \cellcolor{cellbg0}\color{white}4.0 & \cellcolor{cellbg0}\color{white}7.9 & \cellcolor{cellbg76}\color{white}5.4 \\
    gemma-4-E2B-it & \cellcolor{cellbg0}\color{white}4.0 & \cellcolor{cellbg19}\color{black}5.8 & \cellcolor{cellbg34}\color{black}14.7 & \cellcolor{cellbg46}\color{white}5.8 & \cellcolor{cellbg0}\color{white}4.0 & \cellcolor{cellbg16}\color{black}5.5 & \cellcolor{cellbg3}\color{white}49.0 & \cellcolor{cellbg3}\color{white}49.0 & \cellcolor{cellbg3}\color{white}49.0 & \cellcolor{cellbg26}\color{white}11.0 & \cellcolor{cellbg9}\color{black}28.0 & \cellcolor{cellbg3}\color{white}49.0 \\
    gemma-4-E4B-it & \cellcolor{cellbg0}\color{white}4.0 & \cellcolor{cellbg20}\color{white}5.3 & \cellcolor{cellbg35}\color{black}11.9 & \cellcolor{cellbg0}\color{white}4.0 & \cellcolor{cellbg0}\color{white}4.0 & \cellcolor{cellbg55}\color{black}5.5 & \cellcolor{cellbg3}\color{white}49.0 & \cellcolor{cellbg3}\color{white}49.0 & \cellcolor{cellbg3}\color{white}49.0 & \cellcolor{cellbg3}\color{white}12.0 & \cellcolor{cellbg3}\color{white}49.0 & \cellcolor{cellbg3}\color{white}49.0 \\
    gpt-4o & \cellcolor{cellbg0}\color{white}4.0 & \cellcolor{cellbg8}\color{black}5.7 & \cellcolor{cellbg36}\color{black}10.1 & \cellcolor{cellbg0}\color{white}4.0 & \cellcolor{cellbg0}\color{white}4.0 & \cellcolor{cellbg56}\color{black}5.2 & \cellcolor{cellbg3}\color{white}49.0 & \cellcolor{cellbg3}\color{white}49.0 & \cellcolor{cellbg63}\color{white}15.4 & \cellcolor{cellbg20}\color{white}4.6 & \cellcolor{cellbg17}\color{white}12.0 & \cellcolor{cellbg76}\color{white}5.4 \\
    \bottomrule
      \end{tabular}
  \end{adjustbox}
\end{table}

\definecolor{cellbg0}{RGB}{57,167,88}
\definecolor{cellbg1}{RGB}{187,21,38}
\definecolor{cellbg2}{RGB}{225,68,48}
\definecolor{cellbg3}{RGB}{165,0,38}
\definecolor{cellbg4}{RGB}{169,4,38}
\definecolor{cellbg5}{RGB}{249,147,85}
\definecolor{cellbg6}{RGB}{200,34,39}
\definecolor{cellbg7}{RGB}{219,56,43}
\definecolor{cellbg8}{RGB}{241,249,172}
\definecolor{cellbg9}{RGB}{135,203,103}
\definecolor{cellbg10}{RGB}{0,104,55}
\definecolor{cellbg11}{RGB}{93,185,97}
\definecolor{cellbg12}{RGB}{9,121,64}
\definecolor{cellbg13}{RGB}{69,173,91}
\definecolor{cellbg14}{RGB}{8,119,63}
\definecolor{cellbg15}{RGB}{187,226,120}
\definecolor{cellbg16}{RGB}{107,191,100}
\definecolor{cellbg17}{RGB}{90,183,96}
\definecolor{cellbg18}{RGB}{137,204,103}
\definecolor{cellbg19}{RGB}{245,114,69}
\definecolor{cellbg20}{RGB}{179,13,38}
\definecolor{cellbg21}{RGB}{241,102,64}
\definecolor{cellbg22}{RGB}{250,152,87}
\definecolor{cellbg23}{RGB}{245,117,71}
\definecolor{cellbg24}{RGB}{30,154,81}
\definecolor{cellbg25}{RGB}{254,216,132}
\definecolor{cellbg26}{RGB}{22,145,77}
\definecolor{cellbg27}{RGB}{20,142,75}
\definecolor{cellbg28}{RGB}{45,161,85}
\definecolor{cellbg29}{RGB}{23,147,78}
\definecolor{cellbg30}{RGB}{253,183,104}
\definecolor{cellbg31}{RGB}{254,200,119}
\definecolor{cellbg32}{RGB}{160,214,105}
\definecolor{cellbg33}{RGB}{14,130,69}
\definecolor{cellbg34}{RGB}{255,251,184}
\definecolor{cellbg35}{RGB}{252,170,95}
\definecolor{cellbg36}{RGB}{251,253,186}
\definecolor{cellbg37}{RGB}{254,255,190}
\definecolor{cellbg38}{RGB}{122,198,101}
\definecolor{cellbg39}{RGB}{224,242,149}
\definecolor{cellbg40}{RGB}{254,236,159}
\definecolor{cellbg41}{RGB}{19,140,74}
\definecolor{cellbg42}{RGB}{3,110,58}
\definecolor{cellbg43}{RGB}{244,250,176}
\definecolor{cellbg44}{RGB}{21,144,76}
\definecolor{cellbg45}{RGB}{132,202,102}
\definecolor{cellbg46}{RGB}{102,189,99}
\definecolor{cellbg47}{RGB}{78,177,93}
\definecolor{cellbg48}{RGB}{173,220,111}
\definecolor{cellbg49}{RGB}{183,224,117}
\definecolor{cellbg50}{RGB}{193,229,123}
\definecolor{cellbg51}{RGB}{99,188,98}
\definecolor{cellbg52}{RGB}{72,174,92}
\definecolor{cellbg53}{RGB}{201,232,129}
\definecolor{cellbg54}{RGB}{185,225,118}
\definecolor{cellbg55}{RGB}{66,172,90}
\definecolor{cellbg56}{RGB}{39,159,83}
\definecolor{cellbg57}{RGB}{152,211,104}
\definecolor{cellbg58}{RGB}{228,76,52}
\definecolor{cellbg59}{RGB}{248,140,81}
\definecolor{cellbg60}{RGB}{185,19,38}
\definecolor{cellbg61}{RGB}{210,43,39}
\definecolor{cellbg62}{RGB}{218,240,141}
\definecolor{cellbg63}{RGB}{127,200,102}
\definecolor{cellbg64}{RGB}{179,223,114}
\definecolor{cellbg65}{RGB}{162,215,106}
\definecolor{cellbg66}{RGB}{1,106,56}
\definecolor{cellbg67}{RGB}{142,207,103}
\definecolor{cellbg68}{RGB}{25,151,80}
\definecolor{cellbg69}{RGB}{253,193,113}
\definecolor{cellbg70}{RGB}{191,228,122}
\definecolor{cellbg71}{RGB}{255,247,178}
\definecolor{cellbg72}{RGB}{254,237,161}
\definecolor{cellbg73}{RGB}{110,192,100}
\definecolor{cellbg74}{RGB}{24,149,79}
\definecolor{cellbg75}{RGB}{15,132,70}

\begin{table}[htbp]
  \centering
  \caption{RT80 Of the main experiments. This reflect all the models in every task which pixel is the minimum acceptable pixel. Samller means the models has good visual capability on this tasks with higher than 80\% of accuracy.}
  \label{tab:rt80}
  \begin{adjustbox}{max width=\textwidth}
    \begin{tabular}{lrrrrrrrrrrrr}
    \toprule
    model & LTR & SHP & ANM & BLK & CBL & TXT & CHR & CMP & CNT & BLR & TRD & TCT \\
    \midrule
    GLM-4.6V-Flash & \cellcolor{cellbg0}\color{black}6.3 & \cellcolor{cellbg18}\color{black}12.0 & \cellcolor{cellbg33}\color{white}12.6 & \cellcolor{cellbg47}\color{black}7.4 & \cellcolor{cellbg10}\color{white}4.0 & \cellcolor{cellbg57}\color{black}7.9 & \cellcolor{cellbg10}\color{white}49.0 & \cellcolor{cellbg10}\color{white}49.0 & \cellcolor{cellbg10}\color{white}49.0 & \cellcolor{cellbg68}\color{white}14.4 & \cellcolor{cellbg10}\color{white}49.0 & \cellcolor{cellbg3}\color{white}49.0 \\
    InternVL3\_5-14B-Flash & \cellcolor{cellbg1}\color{white}7.5 & \cellcolor{cellbg19}\color{black}19.3 & \cellcolor{cellbg34}\color{black}30.3 & \cellcolor{cellbg48}\color{black}10.3 & \cellcolor{cellbg3}\color{white}4.6 & \cellcolor{cellbg3}\color{white}9.6 & \cellcolor{cellbg10}\color{white}49.0 & \cellcolor{cellbg10}\color{white}49.0 & \cellcolor{cellbg10}\color{white}49.0 & \cellcolor{cellbg69}\color{black}36.0 & \cellcolor{cellbg10}\color{white}49.0 & \cellcolor{cellbg3}\color{white}49.0 \\
    InternVL3\_5-1B-Flash & \cellcolor{cellbg2}\color{white}7.3 & \cellcolor{cellbg20}\color{white}21.7 & \cellcolor{cellbg35}\color{black}37.6 & \cellcolor{cellbg49}\color{black}10.6 & \cellcolor{cellbg10}\color{white}4.0 & \cellcolor{cellbg58}\color{black}9.3 & \cellcolor{cellbg10}\color{white}49.0 & \cellcolor{cellbg10}\color{white}49.0 & \cellcolor{cellbg10}\color{white}49.0 & \cellcolor{cellbg3}\color{white}49.0 & \cellcolor{cellbg10}\color{white}49.0 & \cellcolor{cellbg3}\color{white}49.0 \\
    InternVL3\_5-2B-Flash & \cellcolor{cellbg3}\color{white}7.5 & \cellcolor{cellbg21}\color{black}19.6 & \cellcolor{cellbg36}\color{black}29.3 & \cellcolor{cellbg50}\color{black}11.1 & \cellcolor{cellbg10}\color{white}4.0 & \cellcolor{cellbg35}\color{black}8.9 & \cellcolor{cellbg10}\color{white}49.0 & \cellcolor{cellbg10}\color{white}49.0 & \cellcolor{cellbg10}\color{white}49.0 & \cellcolor{cellbg3}\color{white}49.0 & \cellcolor{cellbg10}\color{white}49.0 & \cellcolor{cellbg3}\color{white}49.0 \\
    InternVL3\_5-30B-A3B-Flash & \cellcolor{cellbg4}\color{white}7.5 & \cellcolor{cellbg22}\color{black}18.5 & \cellcolor{cellbg37}\color{black}29.6 & \cellcolor{cellbg51}\color{black}8.0 & \cellcolor{cellbg10}\color{white}4.0 & \cellcolor{cellbg20}\color{white}9.6 & \cellcolor{cellbg10}\color{white}49.0 & \cellcolor{cellbg10}\color{white}49.0 & \cellcolor{cellbg10}\color{white}49.0 & \cellcolor{cellbg48}\color{black}22.7 & \cellcolor{cellbg10}\color{white}49.0 & \cellcolor{cellbg3}\color{white}49.0 \\
    InternVL3\_5-38B-Flash & \cellcolor{cellbg5}\color{black}7.1 & \cellcolor{cellbg23}\color{black}19.2 & \cellcolor{cellbg38}\color{black}19.4 & \cellcolor{cellbg52}\color{black}7.2 & \cellcolor{cellbg10}\color{white}4.0 & \cellcolor{cellbg59}\color{black}9.0 & \cellcolor{cellbg10}\color{white}49.0 & \cellcolor{cellbg10}\color{white}49.0 & \cellcolor{cellbg10}\color{white}49.0 & \cellcolor{cellbg70}\color{black}24.0 & \cellcolor{cellbg10}\color{white}49.0 & \cellcolor{cellbg3}\color{white}49.0 \\
    InternVL3\_5-4B-Flash & \cellcolor{cellbg6}\color{white}7.4 & \cellcolor{cellbg3}\color{white}22.1 & \cellcolor{cellbg39}\color{black}26.7 & \cellcolor{cellbg53}\color{black}11.4 & \cellcolor{cellbg10}\color{white}4.0 & \cellcolor{cellbg60}\color{white}9.5 & \cellcolor{cellbg10}\color{white}49.0 & \cellcolor{cellbg10}\color{white}49.0 & \cellcolor{cellbg10}\color{white}49.0 & \cellcolor{cellbg71}\color{black}30.7 & \cellcolor{cellbg10}\color{white}49.0 & \cellcolor{cellbg3}\color{white}49.0 \\
    InternVL3\_5-8B-Flash & \cellcolor{cellbg7}\color{white}7.4 & \cellcolor{cellbg4}\color{white}22.0 & \cellcolor{cellbg40}\color{black}32.0 & \cellcolor{cellbg54}\color{black}10.8 & \cellcolor{cellbg10}\color{white}4.0 & \cellcolor{cellbg61}\color{white}9.4 & \cellcolor{cellbg10}\color{white}49.0 & \cellcolor{cellbg10}\color{white}49.0 & \cellcolor{cellbg10}\color{white}49.0 & \cellcolor{cellbg72}\color{black}32.0 & \cellcolor{cellbg10}\color{white}49.0 & \cellcolor{cellbg3}\color{white}49.0 \\
    Qwen3-VL-2B-Instruct & \cellcolor{cellbg8}\color{black}6.7 & \cellcolor{cellbg24}\color{white}10.0 & \cellcolor{cellbg41}\color{white}13.4 & \cellcolor{cellbg3}\color{white}24.0 & \cellcolor{cellbg10}\color{white}4.0 & \cellcolor{cellbg62}\color{black}8.2 & \cellcolor{cellbg10}\color{white}49.0 & \cellcolor{cellbg10}\color{white}49.0 & \cellcolor{cellbg10}\color{white}49.0 & \cellcolor{cellbg3}\color{white}49.0 & \cellcolor{cellbg10}\color{white}49.0 & \cellcolor{cellbg3}\color{white}49.0 \\
    Qwen3-VL-30B-A3B-Instruct & \cellcolor{cellbg9}\color{black}6.4 & \cellcolor{cellbg25}\color{black}16.9 & \cellcolor{cellbg3}\color{white}49.0 & \cellcolor{cellbg55}\color{black}7.1 & \cellcolor{cellbg10}\color{white}4.0 & \cellcolor{cellbg63}\color{black}7.8 & \cellcolor{cellbg10}\color{white}49.0 & \cellcolor{cellbg10}\color{white}49.0 & \cellcolor{cellbg10}\color{white}49.0 & \cellcolor{cellbg73}\color{black}18.7 & \cellcolor{cellbg10}\color{white}49.0 & \cellcolor{cellbg3}\color{white}49.0 \\
    Qwen3-VL-4B-Instruct & \cellcolor{cellbg10}\color{white}6.1 & \cellcolor{cellbg10}\color{white}8.5 & \cellcolor{cellbg42}\color{white}11.0 & \cellcolor{cellbg13}\color{black}7.2 & \cellcolor{cellbg10}\color{white}4.0 & \cellcolor{cellbg64}\color{black}8.0 & \cellcolor{cellbg10}\color{white}49.0 & \cellcolor{cellbg10}\color{white}49.0 & \cellcolor{cellbg10}\color{white}49.0 & \cellcolor{cellbg74}\color{white}14.2 & \cellcolor{cellbg10}\color{white}49.0 & \cellcolor{cellbg3}\color{white}49.0 \\
    Qwen3-VL-8B-Instruct & \cellcolor{cellbg11}\color{black}6.3 & \cellcolor{cellbg26}\color{white}9.7 & \cellcolor{cellbg43}\color{black}28.6 & \cellcolor{cellbg13}\color{black}7.1 & \cellcolor{cellbg10}\color{white}4.0 & \cellcolor{cellbg65}\color{black}7.9 & \cellcolor{cellbg10}\color{white}49.0 & \cellcolor{cellbg10}\color{white}49.0 & \cellcolor{cellbg10}\color{white}49.0 & \cellcolor{cellbg12}\color{white}12.0 & \cellcolor{cellbg10}\color{white}49.0 & \cellcolor{cellbg3}\color{white}49.0 \\
    gemini-2.5-flash-preview & \cellcolor{cellbg12}\color{white}6.1 & \cellcolor{cellbg27}\color{white}9.6 & \cellcolor{cellbg10}\color{white}10.5 & \cellcolor{cellbg10}\color{white}4.0 & \cellcolor{cellbg10}\color{white}4.0 & \cellcolor{cellbg42}\color{white}7.2 & \cellcolor{cellbg10}\color{white}49.0 & \cellcolor{cellbg10}\color{white}49.0 & \cellcolor{cellbg10}\color{white}49.0 & \cellcolor{cellbg75}\color{white}12.9 & \cellcolor{cellbg10}\color{white}49.0 & \cellcolor{cellbg10}\color{white}7.4 \\
    gemma-4-26B-A4B-it & \cellcolor{cellbg13}\color{black}6.3 & \cellcolor{cellbg28}\color{black}10.3 & \cellcolor{cellbg44}\color{white}13.7 & \cellcolor{cellbg10}\color{white}4.0 & \cellcolor{cellbg10}\color{white}4.0 & \cellcolor{cellbg42}\color{white}7.2 & \cellcolor{cellbg10}\color{white}49.0 & \cellcolor{cellbg10}\color{white}49.0 & \cellcolor{cellbg10}\color{white}49.0 & \cellcolor{cellbg33}\color{white}12.7 & \cellcolor{cellbg10}\color{white}49.0 & \cellcolor{cellbg3}\color{white}49.0 \\
    gemma-4-31B-it & \cellcolor{cellbg14}\color{white}6.1 & \cellcolor{cellbg29}\color{white}9.8 & \cellcolor{cellbg44}\color{white}13.7 & \cellcolor{cellbg10}\color{white}4.0 & \cellcolor{cellbg10}\color{white}4.0 & \cellcolor{cellbg66}\color{white}7.2 & \cellcolor{cellbg10}\color{white}49.0 & \cellcolor{cellbg10}\color{white}49.0 & \cellcolor{cellbg10}\color{white}49.0 & \cellcolor{cellbg10}\color{white}10.5 & \cellcolor{cellbg10}\color{white}49.0 & \cellcolor{cellbg27}\color{white}10.8 \\
    gemma-4-E2B-it & \cellcolor{cellbg15}\color{black}6.6 & \cellcolor{cellbg30}\color{black}17.8 & \cellcolor{cellbg32}\color{black}21.7 & \cellcolor{cellbg52}\color{black}7.2 & \cellcolor{cellbg10}\color{white}4.0 & \cellcolor{cellbg65}\color{black}7.9 & \cellcolor{cellbg10}\color{white}49.0 & \cellcolor{cellbg10}\color{white}49.0 & \cellcolor{cellbg10}\color{white}49.0 & \cellcolor{cellbg3}\color{white}49.0 & \cellcolor{cellbg10}\color{white}49.0 & \cellcolor{cellbg3}\color{white}49.0 \\
    gemma-4-E4B-it & \cellcolor{cellbg16}\color{black}6.4 & \cellcolor{cellbg31}\color{black}17.3 & \cellcolor{cellbg45}\color{black}20.0 & \cellcolor{cellbg56}\color{white}6.4 & \cellcolor{cellbg10}\color{white}4.0 & \cellcolor{cellbg67}\color{black}7.8 & \cellcolor{cellbg10}\color{white}49.0 & \cellcolor{cellbg10}\color{white}49.0 & \cellcolor{cellbg10}\color{white}49.0 & \cellcolor{cellbg3}\color{white}49.0 & \cellcolor{cellbg10}\color{white}49.0 & \cellcolor{cellbg3}\color{white}49.0 \\
    gpt-4o & \cellcolor{cellbg17}\color{black}6.3 & \cellcolor{cellbg32}\color{black}12.5 & \cellcolor{cellbg46}\color{black}18.2 & \cellcolor{cellbg10}\color{white}4.0 & \cellcolor{cellbg10}\color{white}4.0 & \cellcolor{cellbg10}\color{white}7.2 & \cellcolor{cellbg10}\color{white}49.0 & \cellcolor{cellbg10}\color{white}49.0 & \cellcolor{cellbg10}\color{white}49.0 & \cellcolor{cellbg12}\color{white}12.0 & \cellcolor{cellbg10}\color{white}49.0 & \cellcolor{cellbg3}\color{white}49.0 \\
    \bottomrule
    \end{tabular}
  \end{adjustbox}
\end{table}

\definecolor{bestcell}{HTML}{D4EDDA}
\definecolor{secondcell}{HTML}{FFF3CD}
\begin{table*}[htbp]
\centering
\small
\setlength{\tabcolsep}{4pt}
\caption{FineSightBench \textbf{perception-only} leaderboard (greedy decoding). Per-task strict accuracy (\%) over the 6 perception subtasks; summary metrics (AUC, RT50, Hall., Mean, Hall@4px) recomputed on the perception subset. \textbf{Bold} on \colorbox{bestcell}{green} = best per column; \underline{underline} on \colorbox{secondcell}{yellow} = second best.}
\label{tab:finesightbench-leaderboard-perception}
\begin{adjustbox}{max width=\textwidth}
\begin{tabular}{lccccccccccc}
\toprule
\multirow{2}{*}{\textbf{Model}} & \multicolumn{6}{c}{\textbf{Perception}} & \multicolumn{5}{c}{\textbf{Metrics}} \\
\cmidrule(lr){2-7}\cmidrule(lr){8-12}
 & LTR & SHP & ANM & BLK & CBL & TXT & AUC$\uparrow$ & RT50$\downarrow$ & Hall.\,(\%)$\downarrow$ & Mean$\uparrow$ & Hall@4px\,(\%)$\downarrow$ \\
\midrule
Gemini-2.5-Flash & \cellcolor{bestcell}\textbf{93.9} & 86.7 & \cellcolor{bestcell}\textbf{80.6} & 99.7 & 99.6 & 89.4 & \cellcolor{bestcell}\textbf{96.5} & \cellcolor{bestcell}\textbf{4.8} & 3.3 & \cellcolor{bestcell}\textbf{74.8} & 6.7 \\
Gemma-4-31B & \cellcolor{bestcell}\textbf{93.9} & \cellcolor{secondcell}\underline{86.9} & 69.8 & \cellcolor{bestcell}\textbf{100.0} & \cellcolor{bestcell}\textbf{100.0} & \cellcolor{bestcell}\textbf{89.9} & \cellcolor{secondcell}\underline{95.5} & \cellcolor{secondcell}\underline{5.3} & 3.4 & \cellcolor{secondcell}\underline{74.5} & 13.4 \\
Gemma-4-26B-A4B & 93.4 & 85.1 & 69.9 & \cellcolor{bestcell}\textbf{100.0} & 99.6 & \cellcolor{secondcell}\underline{89.7} & 95.3 & 5.4 & 2.1 & 71.4 & 9.1 \\
Qwen3-VL-4B & 93.6 & \cellcolor{bestcell}\textbf{87.0} & \cellcolor{secondcell}\underline{80.1} & 86.1 & 99.1 & 85.6 & 95.3 & 5.4 & 4.7 & 62.3 & 5.0 \\
GPT-4o & 93.0 & 81.6 & 67.4 & 99.9 & \cellcolor{secondcell}\underline{99.7} & 89.1 & 93.8 & 5.5 & 0.0 & 70.3 & 0.3 \\
GLM-4.6V-Flash & 93.4 & 81.0 & 73.0 & 84.9 & 95.3 & 85.1 & 93.4 & 5.9 & 2.9 & 66.9 & 4.0 \\
Gemma-4-E4B & 92.6 & 78.3 & 62.9 & 93.0 & 97.4 & 87.3 & 92.0 & 5.8 & \cellcolor{bestcell}\textbf{0.0} & 56.9 & \cellcolor{bestcell}\textbf{0.0} \\
Qwen3-VL-8B & 92.1 & 83.7 & 57.1 & 86.7 & \cellcolor{secondcell}\underline{99.7} & 86.9 & 91.3 & 6.3 & 5.8 & 60.9 & 7.1 \\
InternVL3.5-38B & 87.9 & 72.6 & 64.7 & 85.6 & 99.3 & 82.7 & 91.0 & 7.1 & 0.0 & 62.6 & \cellcolor{bestcell}\textbf{0.0} \\
Gemma-4-E2B & 91.1 & 76.6 & 60.6 & 86.9 & 96.3 & 87.1 & 90.6 & 6.6 & \cellcolor{bestcell}\textbf{0.0} & 55.3 & \cellcolor{bestcell}\textbf{0.0} \\
Qwen3-VL-2B & 90.6 & 84.9 & 72.4 & 41.6 & 92.0 & 86.0 & 86.7 & 8.0 & \cellcolor{bestcell}\textbf{0.0} & 55.3 & \cellcolor{bestcell}\textbf{0.0} \\
InternVL3.5-2B & 84.3 & 68.9 & 53.3 & 75.0 & 97.6 & 83.1 & 86.6 & 8.7 & \cellcolor{bestcell}\textbf{0.0} & 54.8 & \cellcolor{bestcell}\textbf{0.0} \\
InternVL3.5-30B-A3B & 85.6 & 68.1 & 51.1 & 83.1 & 91.0 & 82.3 & 86.0 & 8.7 & \cellcolor{bestcell}\textbf{0.0} & 58.2 & \cellcolor{bestcell}\textbf{0.0} \\
InternVL3.5-8B & 86.1 & 64.1 & 48.9 & 76.9 & 98.1 & 82.6 & 85.7 & 9.4 & 0.0 & 56.9 & \cellcolor{bestcell}\textbf{0.0} \\
Qwen3-VL-30B-A3B & 86.9 & 71.4 & 43.7 & 88.9 & 98.7 & 87.0 & 85.7 & 6.4 & 8.2 & 57.7 & 20.9 \\
InternVL3.5-14B & 85.3 & 69.7 & 49.4 & 78.9 & 91.0 & 81.9 & 85.4 & 8.9 & 0.0 & 56.1 & \cellcolor{bestcell}\textbf{0.0} \\
InternVL3.5-4B & 86.1 & 65.7 & 50.1 & 73.0 & 93.7 & 82.0 & 85.3 & 9.2 & 0.0 & 53.8 & \cellcolor{bestcell}\textbf{0.0} \\
InternVL3.5-1B & 85.3 & 60.3 & 44.1 & 77.1 & 98.6 & 82.4 & 83.8 & 9.7 & \cellcolor{bestcell}\textbf{0.0} & 51.0 & \cellcolor{bestcell}\textbf{0.0} \\
\bottomrule
\end{tabular}
\end{adjustbox}
\end{table*}

\definecolor{bestcell}{HTML}{D4EDDA}
\definecolor{secondcell}{HTML}{FFF3CD}
\begin{table*}[htbp]
\centering
\small
\setlength{\tabcolsep}{4pt}
\caption{FineSightBench \textbf{reasoning-only} leaderboard (greedy decoding). Per-task strict accuracy (\%) over the 6 reasoning subtasks; summary metrics (AUC, RT50, Hall., Mean, Hall@4px) recomputed on the reasoning subset. \textbf{Bold} on \colorbox{bestcell}{green} = best per column; \underline{underline} on \colorbox{secondcell}{yellow} = second best.}
\label{tab:finesightbench-leaderboard-reasoning}
\begin{adjustbox}{max width=\textwidth}
\begin{tabular}{lccccccccccc}
\toprule
\multirow{2}{*}{\textbf{Model}} & \multicolumn{6}{c}{\textbf{Reasoning}} & \multicolumn{5}{c}{\textbf{Metrics}} \\
\cmidrule(lr){2-7}\cmidrule(lr){8-12}
 & CHR & CMP & CNT & BLR & TRD & TCT & AUC$\uparrow$ & RT50$\downarrow$ & Hall.\,(\%)$\downarrow$ & Mean$\uparrow$ & Hall@4px\,(\%)$\downarrow$ \\
\midrule
Gemma-4-31B & \cellcolor{bestcell}\textbf{46.8} & \cellcolor{bestcell}\textbf{42.1} & \cellcolor{secondcell}\underline{52.0} & \cellcolor{secondcell}\underline{83.0} & \cellcolor{bestcell}\textbf{52.3} & \cellcolor{secondcell}\underline{73.0} & \cellcolor{bestcell}\textbf{62.0} & \cellcolor{bestcell}\textbf{9.2} & 3.4 & \cellcolor{secondcell}\underline{74.5} & 13.4 \\
Gemini-2.5-Flash & 37.1 & 32.3 & 51.1 & 80.7 & \cellcolor{secondcell}\underline{47.0} & \cellcolor{bestcell}\textbf{89.3} & \cellcolor{secondcell}\underline{60.7} & 21.5 & 3.3 & \cellcolor{bestcell}\textbf{74.8} & 6.7 \\
Gemma-4-26B-A4B & \cellcolor{secondcell}\underline{38.6} & 33.0 & \cellcolor{bestcell}\textbf{55.1} & \cellcolor{bestcell}\textbf{83.4} & 43.7 & 57.0 & 56.3 & \cellcolor{secondcell}\underline{16.9} & 2.1 & 71.4 & 9.1 \\
GPT-4o & 34.1 & \cellcolor{secondcell}\underline{34.5} & 48.7 & 77.9 & 44.4 & 64.4 & 54.9 & 22.6 & 0.0 & 70.3 & 0.3 \\
GLM-4.6V-Flash & 24.1 & 20.7 & 51.4 & 81.6 & 39.3 & 60.9 & 51.0 & 23.6 & 2.9 & 66.9 & 4.0 \\
InternVL3.5-38B & 26.1 & 18.8 & 38.3 & 69.6 & 39.1 & 56.3 & 46.1 & 25.7 & 0.0 & 62.6 & \cellcolor{bestcell}\textbf{0.0} \\
InternVL3.5-30B-A3B & 16.6 & 14.5 & 33.6 & 71.4 & 35.7 & 55.4 & 42.0 & 26.7 & \cellcolor{bestcell}\textbf{0.0} & 58.2 & \cellcolor{bestcell}\textbf{0.0} \\
InternVL3.5-8B & 16.4 & 15.9 & 31.2 & 68.2 & 37.4 & 48.3 & 40.4 & 28.4 & 0.0 & 56.9 & \cellcolor{bestcell}\textbf{0.0} \\
Qwen3-VL-8B & 7.3 & 4.8 & 34.6 & 78.8 & 37.7 & 48.3 & 38.3 & 31.9 & 5.8 & 60.9 & 7.1 \\
InternVL3.5-14B & 18.0 & 3.9 & 32.7 & 66.1 & 34.4 & 50.7 & 38.0 & 29.4 & 0.0 & 56.1 & \cellcolor{bestcell}\textbf{0.0} \\
Qwen3-VL-30B-A3B & 13.0 & 5.5 & 25.0 & 76.8 & 39.7 & 46.6 & 36.6 & 29.3 & 8.2 & 57.7 & 20.9 \\
Qwen3-VL-4B & 17.9 & 0.2 & 30.7 & 76.1 & 36.0 & 43.0 & 36.4 & 37.3 & 4.7 & 62.3 & 5.0 \\
InternVL3.5-4B & 11.8 & 9.8 & 28.7 & 70.4 & 33.6 & 32.0 & 33.7 & 38.3 & 0.0 & 53.8 & \cellcolor{bestcell}\textbf{0.0} \\
InternVL3.5-2B & 15.5 & 6.1 & 25.8 & 67.1 & 32.3 & 39.3 & 33.6 & 35.2 & \cellcolor{bestcell}\textbf{0.0} & 54.8 & \cellcolor{bestcell}\textbf{0.0} \\
Qwen3-VL-2B & 16.4 & 13.0 & 20.0 & 68.8 & 33.7 & 37.9 & 33.4 & 41.5 & \cellcolor{bestcell}\textbf{0.0} & 55.3 & \cellcolor{bestcell}\textbf{0.0} \\
Gemma-4-E4B & 20.2 & 16.6 & 30.4 & 53.0 & 34.3 & 6.1 & 29.9 & 42.8 & \cellcolor{bestcell}\textbf{0.0} & 56.9 & \cellcolor{bestcell}\textbf{0.0} \\
Gemma-4-E2B & 18.9 & 1.2 & 28.6 & 57.7 & 34.7 & 12.1 & 27.7 & 39.2 & \cellcolor{bestcell}\textbf{0.0} & 55.3 & \cellcolor{bestcell}\textbf{0.0} \\
InternVL3.5-1B & 0.0 & 0.2 & 20.0 & 56.4 & 28.6 & 46.0 & 26.2 & 35.9 & \cellcolor{bestcell}\textbf{0.0} & 51.0 & \cellcolor{bestcell}\textbf{0.0} \\
\bottomrule
\end{tabular}
\end{adjustbox}
\end{table*}

\clearpage
\newpage

\end{document}